\documentclass[twocolumn]{IEEEtran}
\usepackage{cite}
\usepackage[cmex10]{amsmath}
\usepackage{array}
\usepackage{amssymb}
\usepackage{amsfonts}
\usepackage{subeqnarray}
\usepackage{cases}
\usepackage{mathrsfs}
\usepackage{graphicx}
\usepackage{threeparttable}
\usepackage{acronym}
\usepackage{caption}
\usepackage{subcaption}
\usepackage{dsfont}
\usepackage{url}
\usepackage{color}
\usepackage{pifont}
\usepackage{hyperref}
\hypersetup{hidelinks,
	colorlinks=true,
	allcolors=black,
	pdfstartview=Fit,
	breaklinks=true}

\usepackage[linesnumbered,ruled,vlined]{algorithm2e}
\begin{document}

\title{UMS-VINS: United Monocular-Stereo Features for Visual–Inertial Tightly Coupled Odometry}

\author{\IEEEauthorblockN{Chaoyang~Jiang, Xiaoni~Zheng, Zhe~Jin, and Chengpu Yu}

\thanks{This work is supported by the National Natural Science Foundation of China(No.52002026)(\textit{Corresponding author: Chaoyang Jiang}).

C. Jiang, X. Zheng, and Z. Jin are with the School of Mechanical Engineering, Beijing Institute of Technology, Beijing, China, 100081 (e-mail: cjiang@bit.edu.cn; 3120200441@bit.edu.cn; 3220215054@bit.edu.cn)

C. Yu is with the School of Automation, Beijing Institute of Technology, Beijing, China, 100081 (e-mail: yuchengpu@bit.edu.cn;)

}
}

\maketitle

\begin{abstract}

This paper introduces the united monocular-stereo features into a visual-inertial tightly coupled odometry (UMS-VINS) for robust pose estimation. UMS-VINS requires two cameras and a low-cost inertial measurement unit (IMU). The UMS-VINS is an evolution of VINS-FUSION, which modifies the VINS-FUSION from the following three perspectives. 1) UMS-VINS extracts and tracks features from the sub-pixel plane to achieve better positions of the features. 2) UMS-VINS introduces additional 2-dimensional features from the left and/or right cameras. 3) If the visual initialization fails, the IMU propagation is directly used for pose estimation, and if the visual-IMU alignment fails, UMS-VINS estimates the pose via the visual odometry. The performances on both public datasets and new real-world experiments indicate that the proposed UMS-VINS outperforms the VINS-FUSION from the perspective of localization accuracy, localization robustness, and environmental adaptability.   

\end{abstract}

\IEEEpeerreviewmaketitle

\section{Introduction}

Visual odometry (VO) is popular for the pose estimation of robots and unmanned vehicles \cite{1,3}, which is usually categorized into direct and indirect methods. Direct methods minimize the photometric error based on the assumption of photometric invariance, such as DSO \cite{4} and LSD-SLAM \cite{5}. In contrast, indirect methods match features with geometric constraints, such as OV2SLAM \cite{3} and ORB-SLAM \cite{7}. The former is sensitive to the illumination change, and the latter is unsuitable for textureless environments. Inertial measurement unit (IMU)  can somehow remedy VO for the cases of illumination change and/or texture missing \cite{9,10, 12}. Hence, visual-inertial odometry (VIO) has increasingly attracted attention in recent years.

Loosely coupled VIO fuses the integration of IMU observations and the estimation results of VO to achieve better pose estimations, such as SSF \cite{14}. Tightly coupled VIO directly fuses the visual and IMU measurements, such as DVIO \cite{16}, ORB-SLAM3 \cite{7} and VINS-MONO \cite{9}, which is considered outperforming loosely coupled VIO. The illumination change is a great challenge; therefore, the indirect method based VO and the inertial measurements are tightly coupled  for pose estimation is gradually becoming the mainstream of VIO research. However, it still suffers from less robust initialization and sparse features. 

\subsection{Initialization}
The initialization of VIO can be classified into joint initialization and disjoint initialization. The joint one combines the IMU and camera measurements for initialization while the disjoint method separately initializes VO and IMU.  

Martinelli \cite{17} firstly proposed the joint method that provided a closed-form solution for the estimation of the gravity, the accelerometer bias, the initial velocity, and the depth of the visual feature by minimizing the 3D errors of all space points. Kaiser et al. \cite{18} extended this method to estimate the gyroscope bias. They both assumed that the camera pose could be estimated from the IMU preintegration, resulting in unpredictable errors in uncertain environments. Different from \cite{17,18} optimizing the errors of all 3D space points, Shen et al.\cite{19} estimated the visual scale by optimizing the reprojection errors without any prior information. Yang et al. \cite{20} further applied a probabilistic optimization-based initialization procedure to estimate the camera-IMU transformation matrix, the initial velocity, the attitude, and the visual scale. However, all the joint methods applied the linear estimator and assumed reliable IMU preintegration results, leading to less robust initialization in uncertain outdoor environments. 

The disjoint method was firstly proposed in  ORBSLAM-VI \cite{21} in which the vision-only pose estimation is assumed better than the IMU, and the initial values are stepwise estimated by a five-step method. Huang et al. \cite{22} applied a back-propagation mechanism and a global optimization method to initialize IMU, camera external parameters, and visual scales via a three-step procedure. These works \cite{21,22} are effective but time-consuming due to sequentially estimating the gravity direction and accelerometer bias. With prior maps, Zũniga-Nöel et al. \cite{25} provided an analytical solution for the accelerometer bias, the gravity direction, and the visual scale, which reduces the computational cost and improves the initialization performance. VINS series \cite{9,11,24} assume the accelerometer to be unbiased and speed up the initialization from the three-step procedure: 1) compute the gyroscope bias by minimizing the error of relative rotation between two keyframes; 2) estimate the initial velocity, the gravity vector, and the visual scale; and 3) refine the gravity from gravity constraints. These methods \cite{9,11,24, 25} are robust with lower computational costs, but only provide local optimal solutions. ORB-SLAM3 \cite{7,23} provides another three-step procedure with only one step to initialize IMU, making all estimations jointly consistent. However, on the one hand, ORB-SLAM3 \cite{7} requires excellent initial guesses for all parameters; on the other hand, it requires that IMU is fully motivated before the initialization.

Literature on the initialization of tightly coupled VIO is rich, but few focus on the initialization of the tightly coupled visual-inertial odometry when the IMU is not fully motivated. 

\subsection{Front end of the indirect VIO}

Features have great influence on localization accuracy, which can be improved from three perspectives: 1)select proper features for a particular application, 2) increase the number and variety of features, and 3) introduce constraints to enhance the feature tracking. 

Features are usually extracted via FAST \cite{26}, ORB \cite{27}, Harris \cite{35}, and so on.
Sun et al. \cite{30} extracted FAST features and tracked them via the  Kanade-Lucas-Tomasi (KLT) method \cite{32}. The FAST feature reduces the computational cost, but has no anti-noise ability, no photometric invariance, and no rotational invariance. ORB features \cite{7,21} consist of the Oriented FAST feature points \cite{26} and the Rotated BRIEF descriptors \cite{41}. The features are uniformly distributed via quadtree and recharged multiple times for different odometry. Compared with FAST features, ORB features can better balance the robustness and complexity of VIO, but have more requirements for the environments \cite{27}. OKVIS \cite{34} extracted Harris features \cite{35} and applied the uniform distribution to guarantee the robustness by gradually ignoring the weak corners around a strong corner. However, these methods can not avoid the existing disadvantages of features. 

Yu et al. \cite{8} proposed an IMU-aided feature-based method to improve the efficiency of features matching on the assumption that the camera and IMU are strictly aligned. Chen et al. \cite{33} combined the forward and backward optical flow method to track, and added a circle-matched process between stereo camera features to enhance the accuracy and robustness of FAST features tracking. However, the more constraints are added, the fewer are the features, especially in textureless environments. Liu et al. \cite{15} proposed three Kalman filters working on different fusion intervals based on stereo cameras and IMU, which could provide robust solutions in the texture-less environment. However, these methods can not guarantee enough strong features in the environment with the weak texture constraints.

The state-of-the-art methods try to avoid tracking failure by applying more robust features or introducing stronger constraints. However, it is still a great challenge to guarantee enough high-quality features in textureless situations.

\subsection{Contribution}

We focus on the above-mentioned two problems and apply the united monocular and stereo features for tightly-coupled VIO, which we call `UMS-VINS'.  The proposed robust initialization supports the mode switch between VIO and VO. The proposed method can guarantee a successful initialization, even for the cases of degrees of freedom degeneracy. Both 2D features from the left camera and/or right camera and the 3D features are considered, which significantly increase the number of features and improve the robustness of the VINS-FUSION \cite{11}, especially in textureless environments. 

The main contributions of the proposed UMS-VINS are listed as follows:
\begin{enumerate}
    \item The proposed UMS-VINS guarantees a successful
    initialization for the visual-inertial systems, even when robots are initialized from the stationary state or the motion state with degrees of freedom degeneracy.
    \item Tightly coupled VIO with 3D and 2D features extracted from the sub-pixel image can guarantee more high-quality features even in textureless environments, which dramatically improves the robustness of VIO.
    \item Both public datasets and new real-world experiments indicate that the proposed UMS-VINS outperforms the VINS-FUSION from the perspective of localization accuracy, robustness, and environmental adaptability.
\end{enumerate}

\subsection{Outline and Notations}
The rest of the paper is organized as follows. Section II overviews the UMS-VINS. Section III details the proposed algorithm. Section IV presents the experimental results, and the conclusions are given in Section V.

This paper uses the following notations: The subscript $(\cdot)_{c}, (\cdot)_{b}, (\cdot)_{w}$ indicate the camera, the IMU (i.e., the body) and the world coordinates, respectively. $P^{w}_l\in \mathbb{R}^3$ represents the $l$-th map point in the world coordinate. $t^{i}_{j}$, $R^{i}_{j}$, $T^{i}_{j}$, and $q^{i}_{j}$ represent the translation vector, the rotation matrix, the transformation, and the quaternion from coordinate $\it{j}$ to coordinate $\it{i}$, respectively. $\pi : \mathbb{R}^3 \rightarrow \mathbb{R}^2$ denotes the camera projection model. $g^{w}$ is the gravity in the world coordinate.

\section{OVERVIEW}
\begin{figure*}[hbt]
    \centering
    \includegraphics[width=0.89\textwidth]{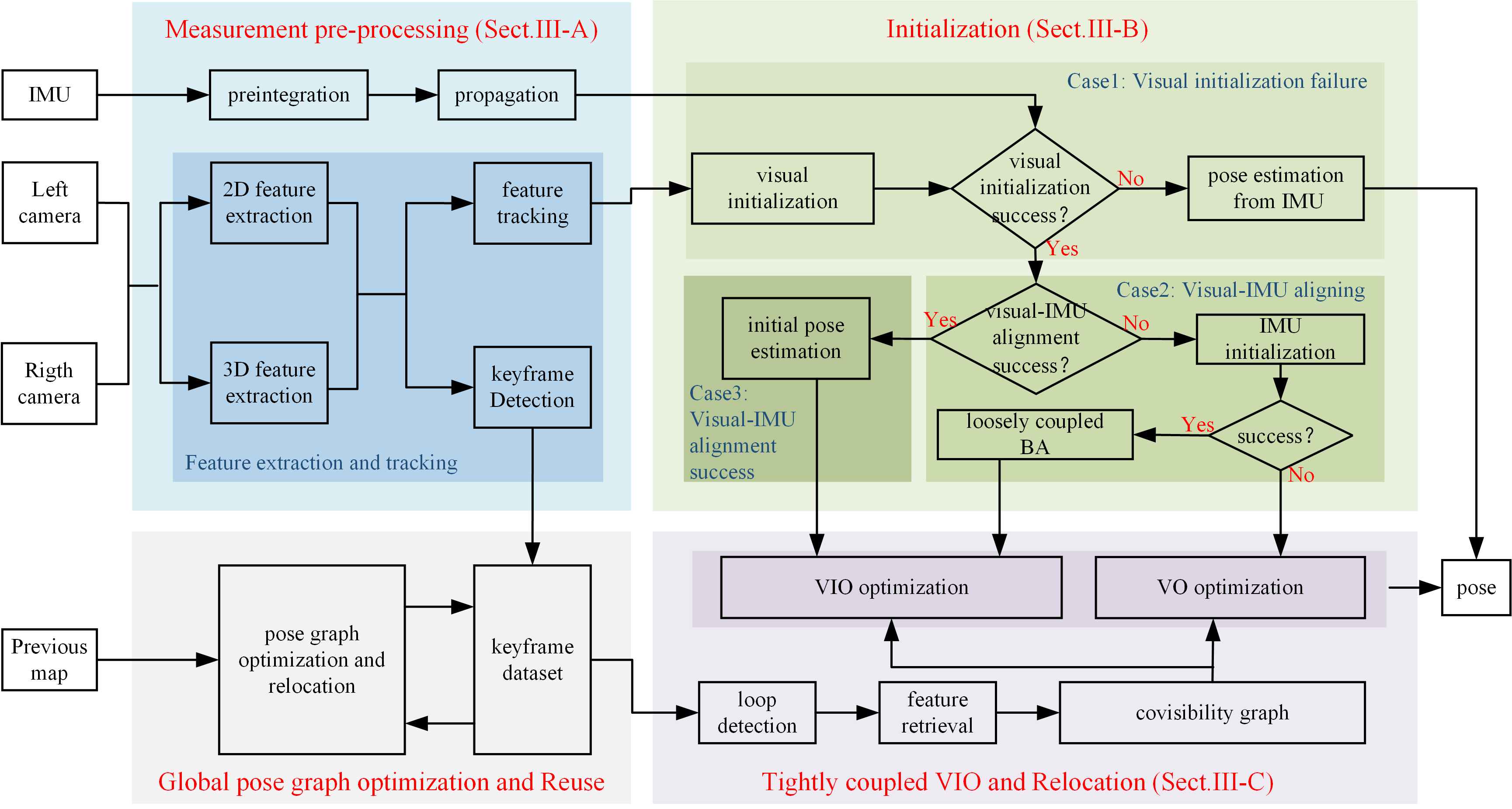}
    \caption{The architecture of the UMS-VINS, which includes four parts: 1)measurements pre-processing, 2) initialization, 3) tightly coupled VIO and relocation, and 4) global pose graph optimization and reuse.}
    \label{Fig:architecture}
\end{figure*}

UMS-VINS applies stereo cameras and an IMU for the pose estimation. The architecture of UMS-VINS is shown in Fig. \ref{Fig:architecture}. Its main components are summarized as follows.

Measurements pre-processing (Section III-A): Firstly, we transform the images from the left and right cameras into sub-pixel images. We then extract both 2D and 3D features from the sub-pixel images, which guarantee enough features. All features are tracked by KLT \cite{32}. In addition, the IMU preintegration between two image frames and the keyframe detection is processed in this stage. 

Initialization(Section III-B): If the visual initialization fails, IMU propagation directly provides the results of the pose estimation. If the visual initialization succeeds but the visual-IMU alignment fails, we estimate poses from VO. If the visual-IMU alignment succeeds, we estimate poses from VIO.  

Tightly coupled VIO and relocation(Section III-C): We apply a bundle adjustment(BA) to real-time optimize the pose from the initial pose estimation with the constraints of VO and IMU. The states of all keyframes in the covisibility graph of the current keyframe are real-time updated. When a loop is detected, the loop-keyframe is applied for optimization, greatly reducing the global translation and orientation drift. The updated states are real-time stored on the map. If the localization fails, UMS-VINS can relocate by map matching.

Global pose map optimization and reuse: The global optimization is implemented to reduce the drift by reusing the available maps, which is the same as that in VINS-FUSION \cite{11} and the details are ignored in this paper. 

\section{Method}
\subsection{Measurement pre-processing}
\begin{figure}[hbt]
    \centering
    \includegraphics[width=0.45\textwidth]{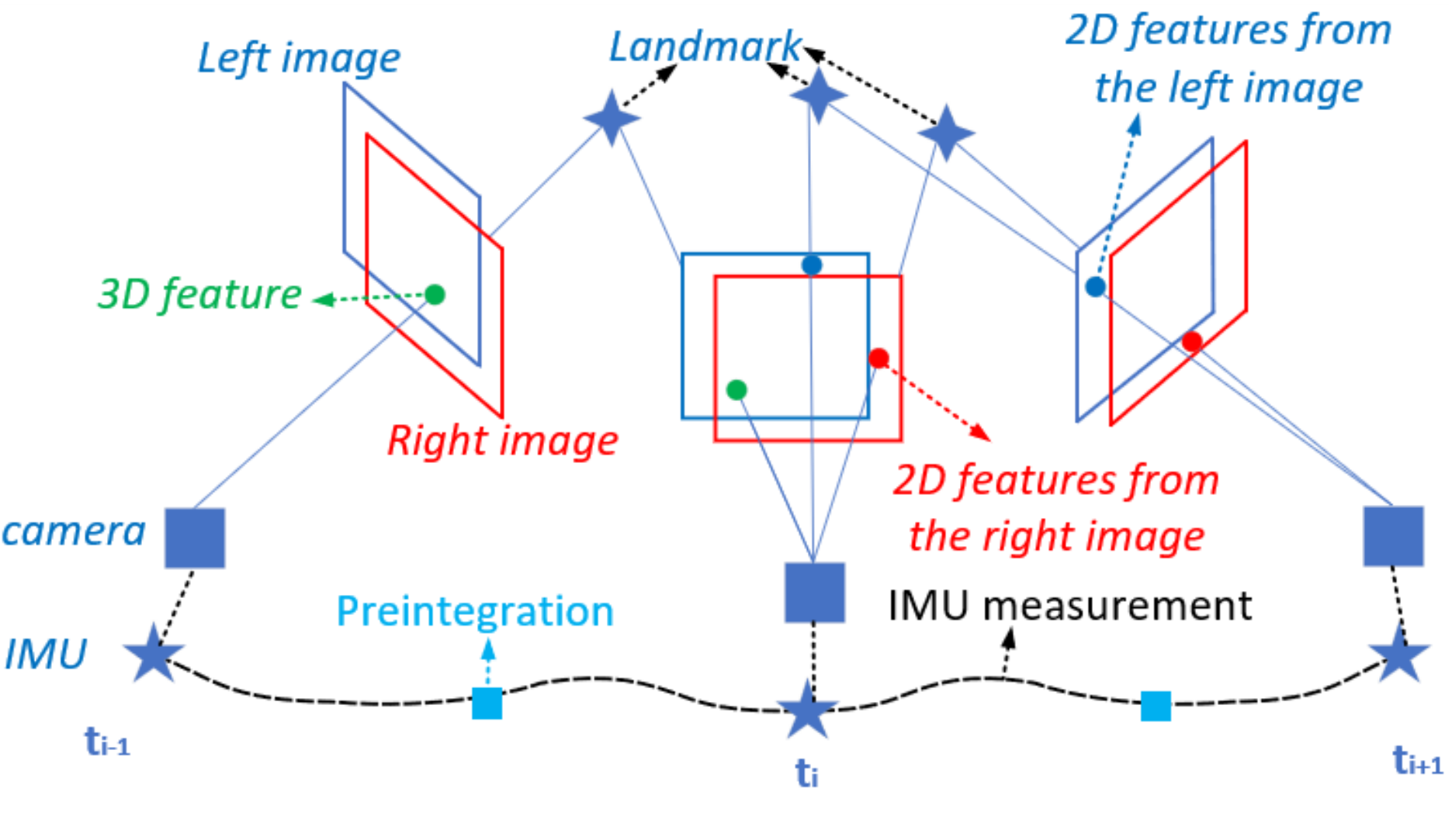}
    \caption{The measurement pre-processing.}
    \label{Fig:pre-processing.}
\end{figure}

Measurement pre-processing is illustrated in Fig. \ref{Fig:pre-processing.}, which includes the image pre-processing and the IMU pre-processing. The features  extracted only from the left or right camera are 2D features. The features extracted from both the left and right cameras are the 3D features. Between two images frames, the features are matched, and the IMU measurements are preintegrated for propagation. After the feature extraction, the keyframe can be detected based on the features. 

\subsubsection{Image pre-processing}
\begin{figure}[hbt]
    \centering
    \includegraphics[width=0.5\textwidth]{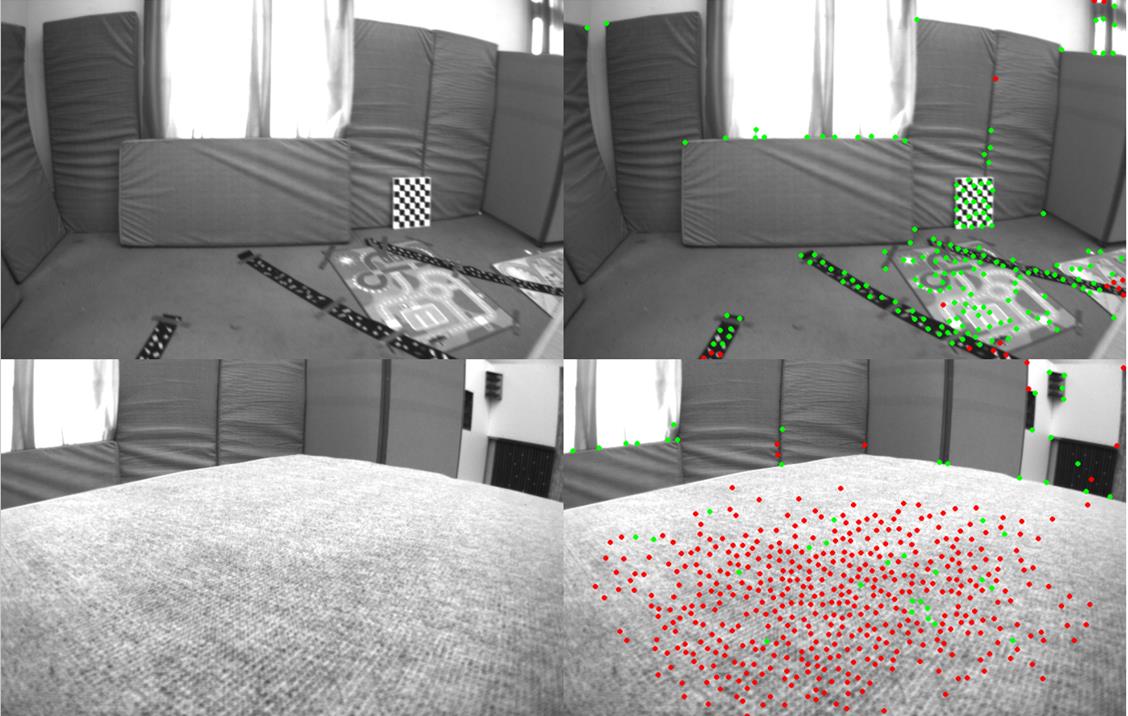}
    \caption{An illustration for 2D and 3D features. Red points represent the 2D features from the left camera. Green points represent the 3D features.}
    \label{Fig:Image pre-processing.}
\end{figure}

We extract the Shi-Tomasi features \cite{35} from the left and the right sub-pixel images. Compared with the traditional pixel plane, features extraction from  sub-pixel planes can obtain more accurate feature positions. The 2D and 3D features are illustrated in Fig. \ref{Fig:Image pre-processing.}. The number of features is fixed. But the ratio of 2D to 3D features varies with the texture of the scene. The number of 3D features is higher in high-texture scenes, whereas the number of 2D features is higher.

We then apply the KLT \cite{32} with the pyramid to track the 2D features (without the depth of the 3D positions) and the 3D features. Here, the 3D features are stereo matching features with the depth of the 3D positions. To balance the time cost and the accuracy, UMS-VINS tracks 2D features with the four-level pyramid and 3D features with the two-level pyramid. If the tracking of 3D features fails, we reuse the four-level pyramid for 3D feature tracking. The outliers are filtered by RANSAC based on epipolar constraints to achieve better feature matching pairs. 

With the extracted features we can proceed the keyframe detection. A frame will be selected as the new keyframe if one of the following three thresholds are triggered:

\begin{enumerate}
\item The time interval between the current frame and the previous keyframe is beyond a pre-defined threshold.
\item The average parallax of features between the current frame and the previous keyframe exceeds a pre-defined threshold.
\item The summation of the number of the 2D features whose depth are initialized by 3D features and the number of 3D features is less than a pre-defined threshold.
\end{enumerate}
More details about the three thresholds can be found from the VINS-MONO \cite{9}.

\subsubsection{IMU preintegration and propagation}
By integrating the raw observations of the accelerometers and gyroscopes during the time slot of two image frames, we can obtain the relative pose of the robot between the two image frames. Further considering the pose and velocity of the last frame, with a propagation we can obtain the pose and velocity of each current frame. More details about IMU preintegration can be found from \cite{25}.

In real applications, the IMU observations easily suffer from saw-tooth waves and impulses when the robots are in some special motion states like swerving and other degrees of freedom degeneracy cases, which leads to severe localization error. Hence, we add a saturation element for the change of the IMU raw observations before the preintegration. 

\subsection{Initialization}
The failure of VIO initialization is mainly because 1) visual initialization failure, and/or 2) the visual-IMU alignment failure. To guarantee the robustness of pose estimation, we divide the initialization into three cases: 1) visual initialization failure, 2) visual-IMU aligning, and 3) visual-IMU alignment success. We show the details of  VIO initialization in case 2, while case 1 and case 3 are its compensations when the VIO initialization fails. Hence, the time cost of initialization is determined by the case 2.

\subsubsection{Visual initialization failure (case 1)}
With the results of measurement pre-processing, we can obtain roughly pose estimation from the IMU preintegration and propagation, and start the visual initialization. 

We apply the MLPnP \cite{40} to estimate the initial poses and feature positions from camera observations. Both 2D and 3D features are tracked, but only the 3D features are used for the visual initialization. 
The initial pose and the inverse depth of 3D features are optimized by minimizing the reprojection error via BA.

The visual initialization tends to fail if the robot starts in textureless environments in which the features are probably not enough and the rate of features mismatch is high. If visual initialization fails, the robot pose is estimated directly by the IMU propagation.

\subsubsection{Visual-IMU aligning (case 2)}

If the visual initialization succeeds, the IMU is aligned with the visual odometry. The procedure includes two steps: IMU initialization and loosely coupled optimization.

\ding{172} IMU initialization: The relative rotation between two image frames equals the corresponding relative rotation in IMU preintegration. Thus, like VINS-FUSION \cite{11}, we can calibrate the gyroscope bias via visual constraints. Then we estimate the gravity direction and velocity from the discrete IMU preintegration equation by solving a least-squares problem. If the IMU initialization fails, UMS-VINS ignores the IMU observations and estimates the pose by VO. Otherwise, we apply loosely coupled BA to optimize the initial states.

\ding{173} Loosely coupled BA: The loosely coupled BA is used to optimize poses, the transformation matrix between the left camera and IMU, the pose estimation from IMU propagation, and IMU biases. The optimized parameters 
\begin{equation*} \label{test1}
X = [x_{0},x_{1},...,x_{n},T^{b}_{c},x^{b}_{0},x^{b}_{1},...,x^{b}_{n},\lambda_{1},\lambda_{2},...,\lambda_{m},b_{a},b_{g}]
\end{equation*}
where $x_k$ is the pose when the $k$-th image keyframe is captured in the left camera coordinate. $T^{b}_{c}$ represents the transformation matrix from the camera coordinate to the IMU coordinate. $x^{b}_{n} = [T^{b_n}_{w},v_n]$ consists of the pose $T^{b_n}_{w}$ and the velocity $v_n$ obtained from IMU propagation. $\lambda_l$ is the inverse depth of the $l$-th feature. $b_a$ and $b_g$ are the biases of the accelerometer and gyroscope, respectively.  

We optimize the parameters $X$ by minimizing the summation of the Mahalanobis norm of all measurement residuals, i.e., 
\begin{eqnarray*}
\min_{X}  \sum\limits_{k\in B} \left[\parallel r_{B_{k}}\parallel_{2} + \parallel r_{BC_{k}}\parallel_{2}\right]
+ \sum\limits_{j\in C} \parallel r_{C_{j}}\parallel_{2}
\end{eqnarray*}
where $r_{B_{k}}$ is the residual of IMU preintegration , $B$ is a set consisting of the indices of all IMU observations. 
\begin{eqnarray} \label{IMUrB}
&&r_{B}(\hat{z}^{b_k}_{b_{k+1}},X)= \left[
            \begin{array}{ccccc}
                e_R\\
                e_p\\
                e_v\\
                e_{b_{a}}\\
                e_{b_{w}}\\
            \end{array}
        \right] \\ \nonumber
&& = \left[
           \begin{array}{ccccc}
               2[{\gamma^{b_k}_{b_{k+1}}}^{-1} \bigotimes q^{w}_{b_k} \bigotimes q^{w}_{b_{k+1}}]_{xyz} \\
               R^{b_k}_{w}(p^{b_{k+1}}_{w} - p^{b_{k}}_{w} - v^{b_{k}}_{w}\Delta t_{k} - \frac{1}{2}g^{w}\Delta t^{2}_{k}) - \alpha^{b_{k}}_{b_{k+1}} \\
               R^{b_k}_{w}(v^{b_{k+1}}_{w} - v^{b_{k}}_{w} + g^{w}\Delta t_{k}) - \beta^{b_{k}}_{b_{k+1}} \\
               b_{a_{k+1}} - b_{a_{k}}\\
               b_{w_{k+1}} - b_{w_{k}}
           \end{array}
        \right]
\end{eqnarray}
where $e_R, e_p, e_v, e_{b_{a}}, e_{b_{w}}$ are the errors about the rotation matrix, position, velocity, and bias of accelerometer and gyroscope, respectively. 
$\alpha^{b_{k+1}}_{b_k}$, $\beta^{b_{k+1}}_{b_k}$, and $\gamma^{b_{k+1}}_{b_k}$ are the pre-integration items, respectively.
\begin{eqnarray*}
\left \{
\begin {aligned}
\alpha^{b_{k+1}}_{b_k} = {\iint_{t_{k}}^{t_{k+1}}}[R^{b_{k}}_{t}(\hat{a}_{t}-b_{a_{t}})]dt^{2} \\
\beta^{b_{k+1}}_{b_k} = {\int_{t_{k}}^{t_{k+1}}}[R^{b_{k}}_{t}(\hat{a}_{t}-b_{a_{t}})]dt \\
\gamma^{b_{k+1}}_{b_k} = {\int_{t_{k}}^{t_{k+1}}}\frac{1}{2}\Omega(\hat{w}_t - b_{w_{t}})\gamma^{b_{k}}_{t}dt 
\end {aligned}
 \right.
\end{eqnarray*}
The residual for the transformation matrix is
\begin{eqnarray*}
 r_{BC_{k}} &=& {T^{b_k}_{bc}T^{b_{k+1}}_{bc}}^{-1} = (T^{b_k}_{w}{T^{c_{k}}_{w}}^{-1}){(T^{b_{k+1}}_{w}{T^{c_{k+1}}_{w}}^{-1})}^{-1}.
\end{eqnarray*}
The residual of reprojection error calculated from features is
\begin{equation}\label{test}
    r_{C_{j}} = (\hat{P}^{c_j}_l - \frac{P^{c_j}_l}{\parallel P^{c_j}_l\parallel}) 
\end{equation}
where  
\begin{eqnarray} \label{eq:pi}
\hat{P}^{c_j}_l &=& \pi^{-1}_{c}(\left[
\begin{array}{cc}
\hat{u}^{c_j}_l\\
\hat{v}^{c_j}_l
\end{array}
\right]) \\
P^{c_j}_l &=& T^{-1}_{wc_j}T_{wc_i} \frac{1}{\lambda_l}\pi^{-1}_{c}(\left[
\begin{array}{cc}
\hat{u}^{c_i}_l\\
\hat{v}^{c_i}_l
\end{array}
\right]),
\end{eqnarray}
$T^{b_k}_{bc}$ is the transformation matrix from the $k$-th left camera keyframe to the IMU coordinate. $T^{c_{k}}_{w}$ is the transformation matrix from the world coordinate to the $k$-th left camera keyframe. $\hat{P}^{c_j}_l$ obtained from the position in the $j$-th pixel plane of the left camera by the inverse projection $\pi^{-1}_{c}$. 
$\hat{u}^{c_j}_l$ and $\hat{v}^{c_j}_l$ are the pixel position in the $j$-th pixel plane of the left camera. $P^{c_j}_l$ is obtained from the $i$-th pixel plane of the left camera. $C$ is a set consisting of the indices of all 3D features and poses of the left camera keyframes.

With the loosely coupled BA, we can obtain the optimized parameters $X$, which will be used as an input of the monocular-stereo tightly coupled VIO. We will repeat this process once. If the difference between the two results of loosely coupled optimization is within a threshold, the initialization succeeds; otherwise, the visual-IMU alignment is considered unsuccessful. 

\subsubsection{Visual-IMU alignment success (case 3)} If the visual odometry and IMU are successfully aligned, the aligning process is not required, and the initial pose would be estimated by MLPnP \cite{40} with the IMU constraints. The estimated pose then be submitted to tightly coupled VIO for further optimization. 

\subsection{Tightly coupled VIO}
\begin{figure}[hbt]
    \centering
    \includegraphics[width=0.45\textwidth]{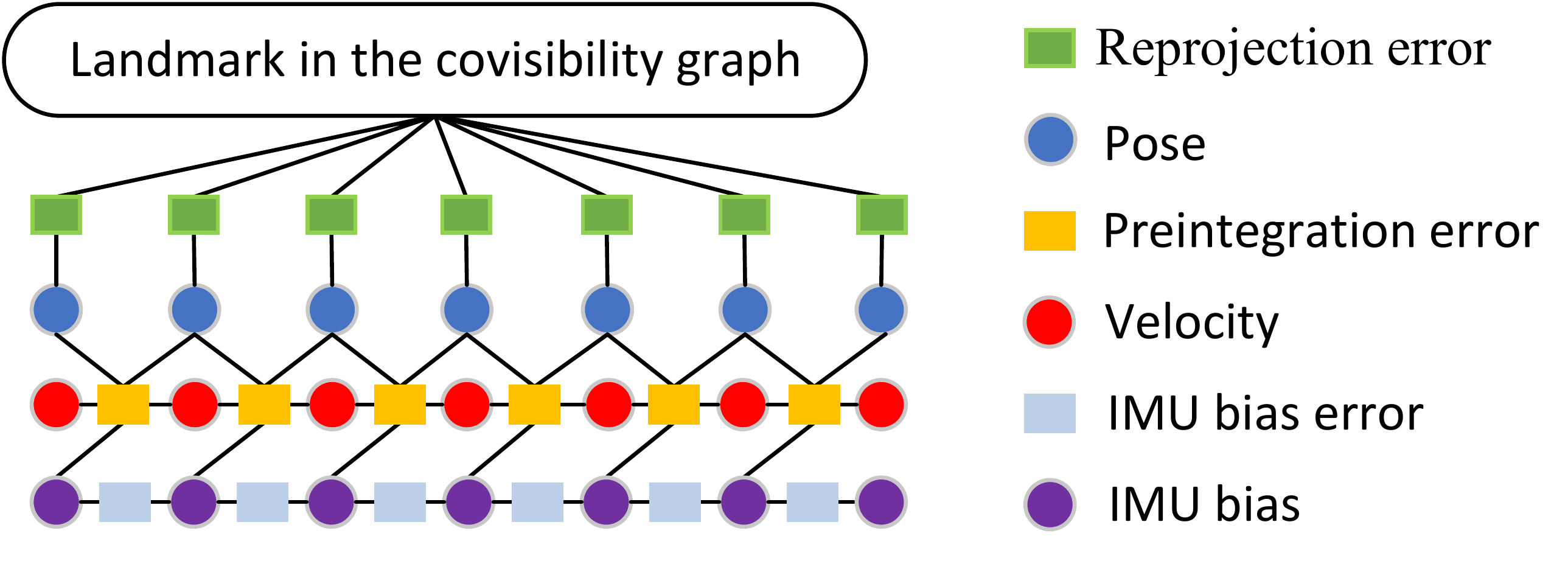}
    \caption{The factor graph for the proposed tightly coupled VIO.}
    \label{Fig: factor graph}
\end{figure}

Two threads simultaneously run in the proposed VIO. In one thread, 3D features are used for pose optimization. In the other thread, 2D features in the left image are added for the pose optimization. The optimization procedure stops if the two results are similar; otherwise, we introduce the 2D features in the right image for further pose optimization. 
An illustration of tightly coupled VIO is shown in Fig. \ref{Fig: factor graph}. The optimized state includes the following parameters:
\begin{equation} \label{eqVIOstate}
        X_{\mathrm{T}} = [x_{0},x_{1},...,x_{n},T^{b}_{c},\lambda_{1},\lambda_{1},...,\lambda_{m}] 
\end{equation}
where
\begin{eqnarray*}
x_k &=& [p^{w}_{b_k},v^{w}_{b_k},q^{w}_{b_k},b_{a},b_{g}],k\in B
\end{eqnarray*}
$x_k$ is the IMU states corresponding to the \emph{k}-th image keyframe, which includes $p^{w}_{b_k}$ representing the position of world coordinate frame relative to the IMU coordinate frame, the corresponding relative velocity $v^{w}_{b_k}$, the relative rotation $q^{w}_{b_k}$, the accelerometer bias $b_a$ and the gyroscope bias $b_g$, all in the IMU coordinate frame. $\lambda_l$ is the inverse depth of the $l$-th feature (2D  or 3D ). We apply BA for pose estimation, i.e., optimizing $X_{\mathrm{T}}$,  in the VIO. 

\subsubsection{Monocular-stereo-inertial optimization for camera-rate state estimation} 
To optimize $X_{\mathrm{T}}$, we minimize the summation of the Mahalanobis norm of all measurement residuals, and yield the following maximum \textit{a posterior} (MAP) estimation problem.
\begin{eqnarray}
 \min_{X_{\mathrm{T}}} \sum\limits_{k\in B} \parallel r_{B_{k}}(X_{\mathrm{T}})\parallel^{2} + \sum\limits_{j\in C} \rho (\parallel r_{C_{j}}(X_{\mathrm{T}})\parallel^{2}) \label{eqMAP}
\end{eqnarray}
where $\rho(\cdot)$ is the Huber norm \cite{40} which is defined as:
\begin{equation*}
    \rho(s) =
    \left\{
    \begin{array}{cc}
        s & s \geq 1\\
        2\sqrt{s} - 1 & s < 1
    \end{array}
    \right.,
\end{equation*}
$r_{B_{k}}(X_{\mathrm{T}})$ is the residual of IMU preintegration. $r_{C_j}(X_{\mathrm{T}}) $ is the residual of visual measurement.

We search all the keyframes in the previous maps to find the covisibility keyframes of the current keyframe. Both 2D features and 3D features are optimized. The residual of the visual odometry includes two parts as shown in \eqref{c(i,j)} : 1) the reprojection residual of the stereo odometry, and 2) the reprojection residual of the monocular odometry in the left and/or right cameras. 
\begin{equation} 
r_{C_j}(X_{\mathrm{T}}) = \beta_{1}(\hat{P}^{c_j}_l - \frac{P^{c_j}_l}{\parallel P^{c_j}_l\parallel}) + \beta_{2}(\hat{P}^{c^{'}_j}_l - \frac{P^{c^{'}_j}_l}{\parallel P^{c^{'}_j}_l\parallel})
\label{c(i,j)}
\end{equation}
where $\beta_{1}$ and $\beta_{2}$ are two weights. Similar with \eqref{eq:pi} and (3),  $\hat{P}^{c_j}_l$ and $\hat{P}^{c^{'}_j}_l$ are respectively obtained from the position of 3D features and 2D features in the $j$-th pixel plane of the left camera by the inverse projection $\pi^{-1}_{c}$.

The 2D features in the left camera and the 3D features are used for residual minimization. The 2D features in the right camera would be involved if 1) the number of 3D features is not enough, or 2) the difference between the poses estimated from the monocular (left camera)-stereo-IMU odometry and the poses estimated from the stereo-IMU odometry exceeds a pre-defined threshold. 

With the solution of \eqref{eqMAP},  we then update the rotation of all keyframes with the relative rotation of the first keyframe in the covisibility graph. If the reprojection error of a feature is larger than a certain threshold after the above-mentioned optimization, the feature is considered to be an outlier. Removing the outliers, we apply the BA again. The second round of BA is to refine the result of the first round. The time cost of the second round optimization is set less than an empirical threshold. The threshold should guarantee the accomplishment of the optimization and the update frequency of the pose estimation unless a higher performance processor is required. To save storage resources, after the second round optimization, UMS-VINS only stores the best 100-300 features. 

\subsubsection{Triangulation} The monocular odometry requires triangulation to obtain 3D maps. Here, we apply triangulation to match both 2D and 3D features. On the one hand, triangulation can help the feature tracking. On the other hand, such triangulation can enable outliers filtering in BA optimization.

\subsubsection{Closed-loop candidate keyframe selection}
We apply the DBoW2 \cite{41} for loop-back detection. If the number of the feature pairs in the current keyframe and a covisibility keyframe exceeds a pre-defined threshold, the covisibility keyframe is considered as a closed-loop candidate keyframe. We only consider the 3D features in the closed-loop detection because the 3D features are considered more informative, and 2D features are firstly discarded in feature retrieval.  

\subsubsection{Tightly coupled relocation}
We implement the closed-loop detection for each new keyframe. If a loop keyframe is found, we update poses to reduce the drift via the pose graph optimization from the loop keyframe to the current keyframe. 

For the closed-loop optimization, a relocation error term is introduced into the original error in \eqref{eqMAP}, and the optimization problem becomes 
\begin{eqnarray*}
\min_{X_{\mathrm{T}}} &&\sum\limits_{k\in B} \parallel r_{B_{k}}(X_{\mathrm{T}})\parallel^{2} + \sum\limits_{j\in C} \rho (\parallel r_{C_j}(X_{\mathrm{T})} \parallel^{2}) \\ 
               &&+ \sum\limits_{(l,v)\in L} \rho(\parallel r_{C_{(l,v)}}(X_{\mathrm{T}})\parallel_{2})
\end{eqnarray*}
where $r_{C_{(l,v)}}(X_{\mathrm{T}})$ represents the reprojection error. $L$ is the set of the features which matched with features in the current keyframe in the loop-closure frames. $(l,v)$ represents the $l$-th feature in the $v$-th closed-loop keyframe.  

We optimize the current keyframe based on the corresponding closed-loop keyframe. The map points are updated via the following correction equation
\begin{equation*}
p^c_{\mathrm{new}} = T^{c_{\mathrm{new}}}_{w} \; T^{w}_{c_v}\; p^c_{\mathrm{o}}
\end{equation*}
where $T^{c_{\mathrm{new}}}_{w}$ is the transformation matrix from the world coordinate to the current camera coordinate. $T^{w}_{c_v}$ represents the transformation matrix from the camera candidate corresponding to the close-loop keyframe to the world coordinate. $p^c_{\mathrm{o}}$ is the original map point in the camera coordinate, and $p^c_{\mathrm{new}}$ is the updated map point in the camera coordinate.

The Ceres Solver \cite{31} is applied to solve the MAP problem \eqref{eqMAP} to optimize the VIO state $X_\mathrm{T}$. More details about the Ceres Solver for MAP problem can be found from \cite{9,11,34}.

\section{Experimental results}
 We compare the proposed UMS-VINS with VINS-FUSION \cite{11} and ORB-SLAM3 \cite{7} in Stereo-Inertial mode in public datasets. Then we compare the proposed UMS-VINS with VINS-FUSION \cite{11} in the new real-world experiments to verify the effectiveness. The algorithm runs on an Intel Core I7-8700 CPU computer. The root-mean-square error (RMSE) of the absolute trajectory error (ATE), i.e., the RMSE of the trajectory, is used as the criterion to assess the algorithms mentioned above. We apply the open-source package EVO provided by the EuRoC public dataset \cite{42} to calculate the trajectory error.

\subsection{Public Dataset}

The public dataset EuRoC is collected by a micro-aerial vehicle (MAV) in indoor environments. The dataset includes stereo images, synchronized IMU raw data, and the groundtruth. More details about EuRoC can be found in \cite{42}, which has been widely used for the testing of SLAM algorithms. We selected ten sequences from the dataset EuRoC \cite{42}, which contains 11 sequences, for testing. Five sequences were collected in a factory hall, while the other five were collected in a small room. The unused sequence suffers from time stamps missing \cite{3}.
\begin{table}[tbh]
\caption{Comparison in the EuRoC dataset (RMSE in m)}
\begin{center}
\begin{tabular}{cccc}
\hline\hline
& VINS-FUSION & ORB-SLAM3 & UMS-VINS\\
\hline
MH\_01 & 0.24 & 0.04 & 0.04 \\ 

MH\_02 & 0.18 & 0.03 & 0.05\\

MH\_03 & 0.23 & 0.03 & 0.10\\

MH\_04 & 0.39 & 0.06 & 0.09\\

MH\_05 & 0.19 & 0.09 & 0.13\\

V1\_01 & 0.1 & 0.04 & 0.06\\

V1\_02 & 0.1 & 0.01 & 0.08\\

V1\_03 & 0.11 & 0.02 & 0.10\\

V2\_01 & 0.12 & 0.04 & 0.05\\

V2\_02 & 0.1 & 0.01 & 0.05\\
\hline\hline
\end{tabular}
\label{Table-pub}
\end{center}
\end{table}

\begin{figure}[hbt]
\begin{subfigure}[t]{0.5\linewidth}
\centering
\includegraphics[width=1\linewidth]{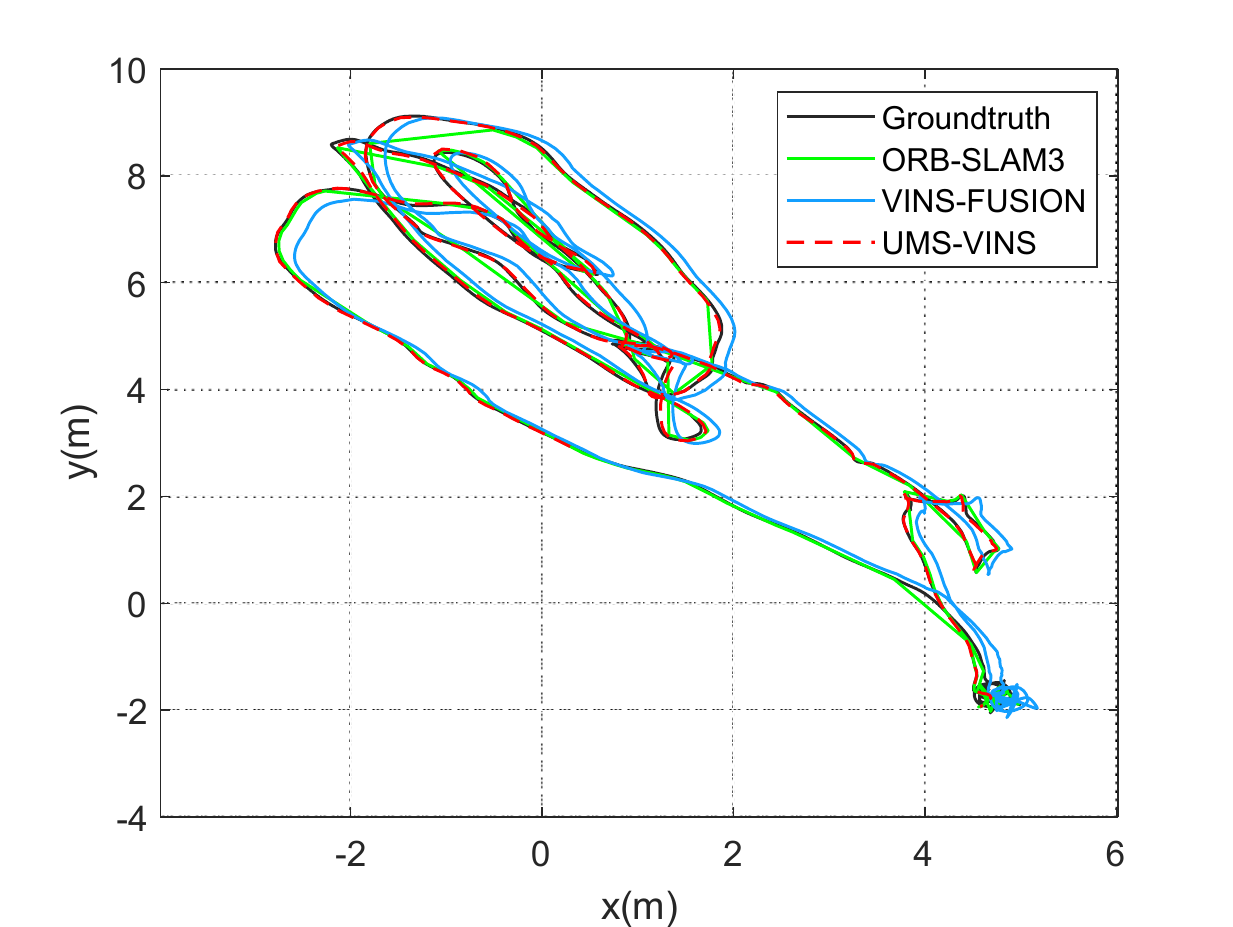}
\caption{Results for MH\_01(factory hall).}
\end{subfigure}%
\begin{subfigure}[t]{0.5\linewidth}
\centering
\includegraphics[width=1\linewidth]{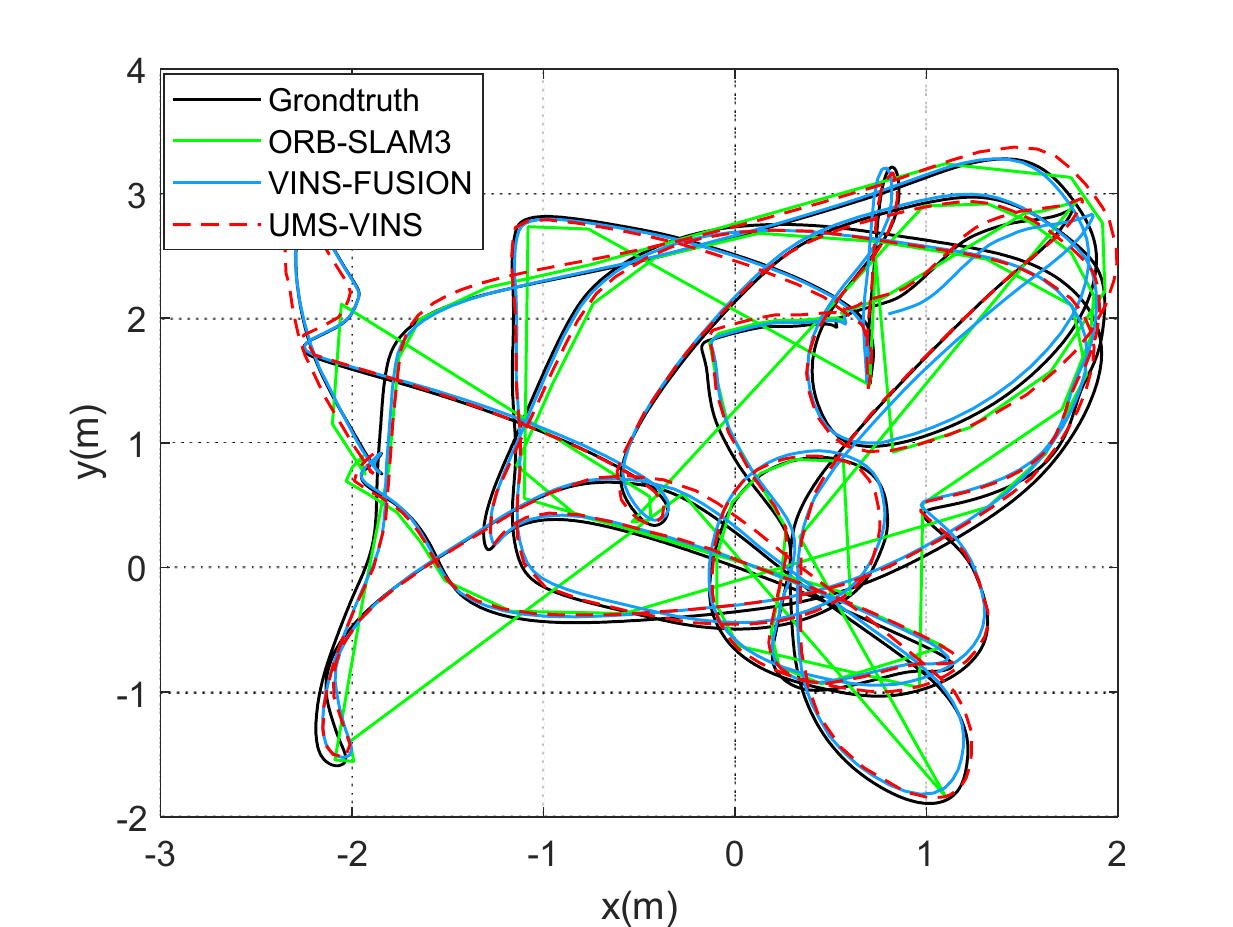}
\caption{Results for V1\_02(small room).}
\end{subfigure}
\begin{subfigure}[t]{0.5\linewidth}
\centering
\includegraphics[width=1\linewidth]{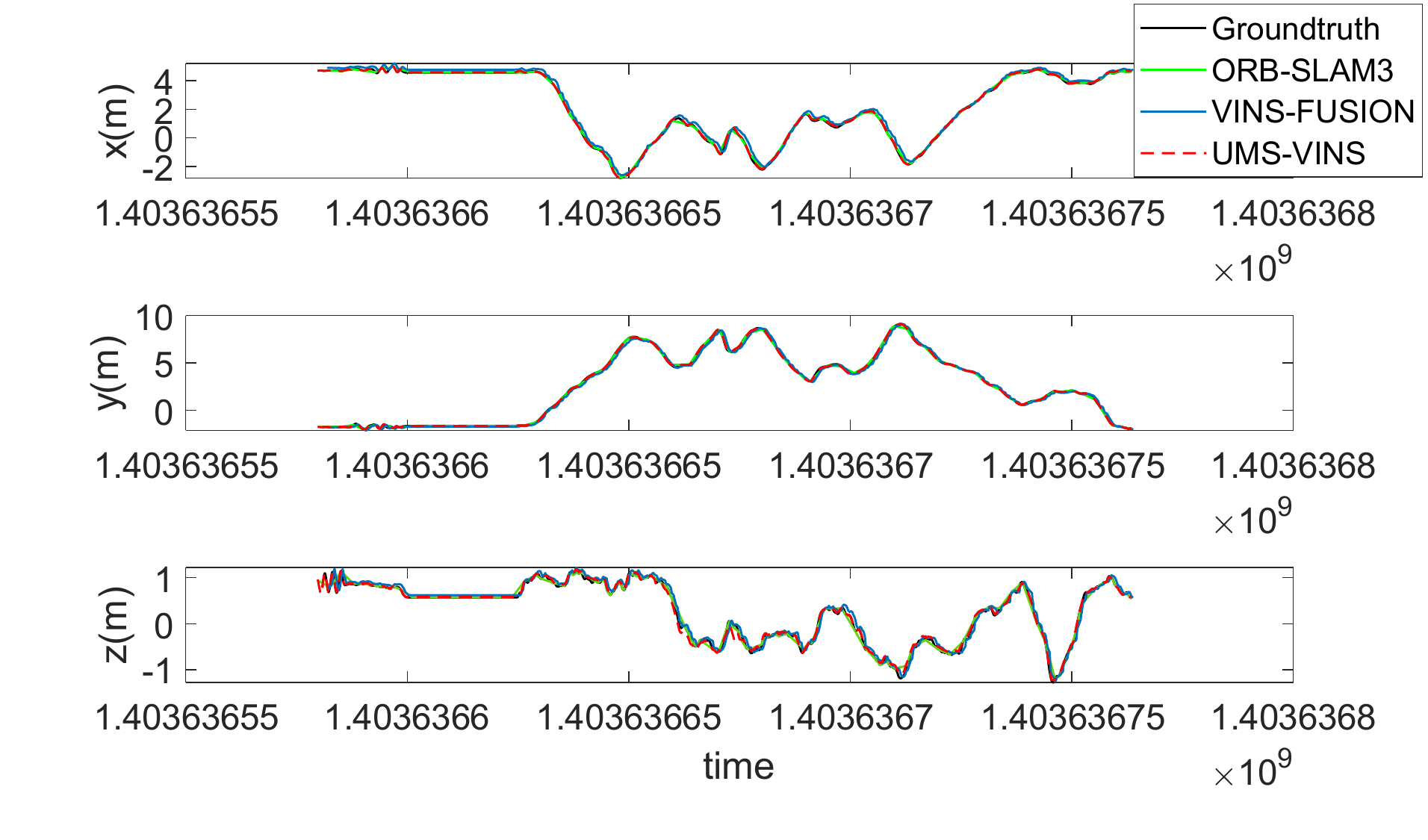}
\caption{Results for MH\_01 from xyz directions.}
\end{subfigure}%
\begin{subfigure}[t]{0.5\linewidth}
\centering
\includegraphics[width=1\linewidth]{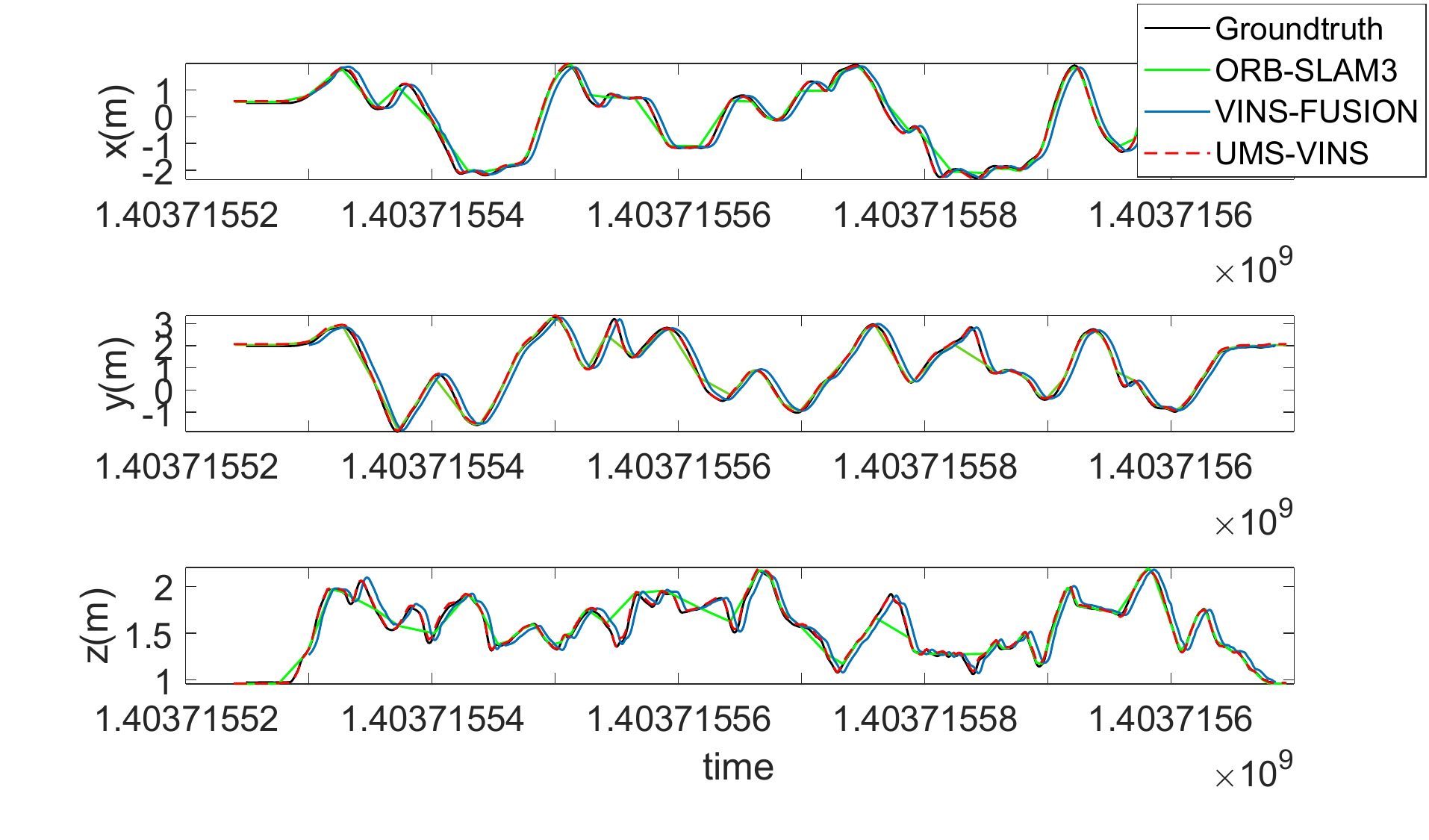}
\caption{Results for V1\_02 from xyz directions.}
\end{subfigure}

\caption{The trajectories estimated from EuRoC datasets using ORB-SLAM3, VINS-FUSION, and UMS-VINS.}
\label{Fig:Trajectories estimated with UMS-VINS and VINS-FUSION on EuRoC datasets.}
\end{figure}
The results are shown in Table \ref{Table-pub}, which indicates that the performances in terms of accuracy are very close to ORB-SLAM3 and outperform VINS-FUSION in all the ten cases. ORB-SLAM3 has high localization accuracy but with lower robustness. We then show the results of MH\_01 (factory hall) and V1\_02 (small room) sequences in Fig. \ref{Fig:Trajectories estimated with UMS-VINS and VINS-FUSION on EuRoC datasets.}.  We can see from  Fig. \ref{Fig:Trajectories estimated with UMS-VINS and VINS-FUSION on EuRoC datasets.} that the trajectory drift of VINS-FUSION is obviously larger than that of the proposed UMS-VINS.

\begin{table*}[tbh]
\caption{Comparison in the EuRoC dataset during 30 times (RMSE in m)}
\begin{center}
\begin{tabular}{cccccccccccc}
\hline\hline
MHO1& & & & & & & & & & & Variance \\
\hline
ORB-SLAM3 & 0.0396 & 0.0374 & 0.0401 & 0.0419 & 0.0432 & 0.0476 & 0.0513 & 0.0385 & 0.0407 & 0.0544 & 0.0042\\
& 0.0407 & 0.0386 & / & 0.0404 & 0.0424 & 0.0415 & 0.0421 & 0.0353 & 0.0375 & 0.0400 \\
& 0.0375 & 0.0462 & 0.0459 & 0.0431 & 0.0417 & 0.0408 & 0.0343 & 0.0406 & 0.0430 & 0.0425 \\ 

VINS-FUSION & 0.1769 & 0.2257 & 0.2318 & 0.2209 & 0.2231 & 0.2173 & 0.2281 & 0.2068 & 0.2193 & 0.2107 & 0.0156\\
& 0.2289 & 0.2180 & 0.2335 & 0.2377 & 0.2299 & 0.2498 & 0.2120 & 0.2331 & 0.2375 & 0.2068 \\
& 0.1934 & 0.2193 & 0.2207 & 0.2433 & 0.2089 & 0.1964 & 0.2178 & 0.2274 & 0.2421 & 0.2335\\

UMS-VINS & 0.0390 & 0.0392 & 0.0380 & 0.0439 & 0.0420 & 0.0452 & 0.0412 & 0.0405 & 0.0415 & 0.0436  & 0.0021\\
& 0.0448 & 0.0443 & 0.0437 & 0.0453 & 0.0407 & 0.0423 & 0.0439 & 0.0418 & 0.0432 & 0.0452 \\
& 0.0424 & 0.0456 & 0.0405 & 0.0458 & 0.0445 & 0.0415 & 0.0436 & 0.0421 & 0.0448 & 0.0433\\

\hline
V102& & & & & & & & & & & Variance \\
\hline

ORB-SLAM3 & 0.0695 & 0.0649 & 0.0663 & 0.0636 & / & 0.0689 & 0.0649 & / & 0.0706 & 0.0629 & 0.0038\\
& 0.0636 & 0.0755 & 0.0704 & / & 0.0631 & 0.0650 & 0.0683 & 0.059 & 0.0636 & 0.0660 \\
& / & 0.0628 & 0.0674 & 0.0640 & 0.0579 & / & 0.0664 & 0.0668 & 0.0655 & 0.0599\\

VINS-FUSION & 0.3042 & 0.3038 & 0.3134 & 0.3038 & 0.3031 & 0.3077 & 0.3069 & 0.3165 & 0.3033 & 0.3019 & 0.0078\\
& 0.3036 & 0.3015 & 0.3025 & 0.3154 & 0.2995 & 0.2966 & 0.2937 & 0.2933 & 0.2895 & 0.2896 \\ 
& 0.2876 & 0.2901 & 0.2919 & 0.2971 & 0.3102 & 0.3029 & 0.2964 & 0.3013 & 0.3082 & 0.3102\\

UMS-VINS & 0.0834 & 0.0922 & 0.0836 & 0.0923 & 0.0914 & 0.0936 & 0.0848 & 0.0846 & 0.0925 & 0.0923 & 0.0032\\
& 0.0912 & 0.0878 & 0.0943 & 0.0939 & 0.0882 & 0.0892 & 0.0884 & 0.0872 & 0.0869 & 0.0913 \\
& 0.0873 & 0.0916 & 0.0907 & 0.0927 & 0.0916 & 0.0937 & 0.0932 & 0.0872 & 0.0931 & 0.0905 \\
\hline\hline
\end{tabular}
\label{Table-init}
\end{center}
\end{table*}

\begin{figure}[hbt]
\begin{subfigure}[t]{0.53\linewidth}
\centering
\includegraphics[width=1\linewidth]{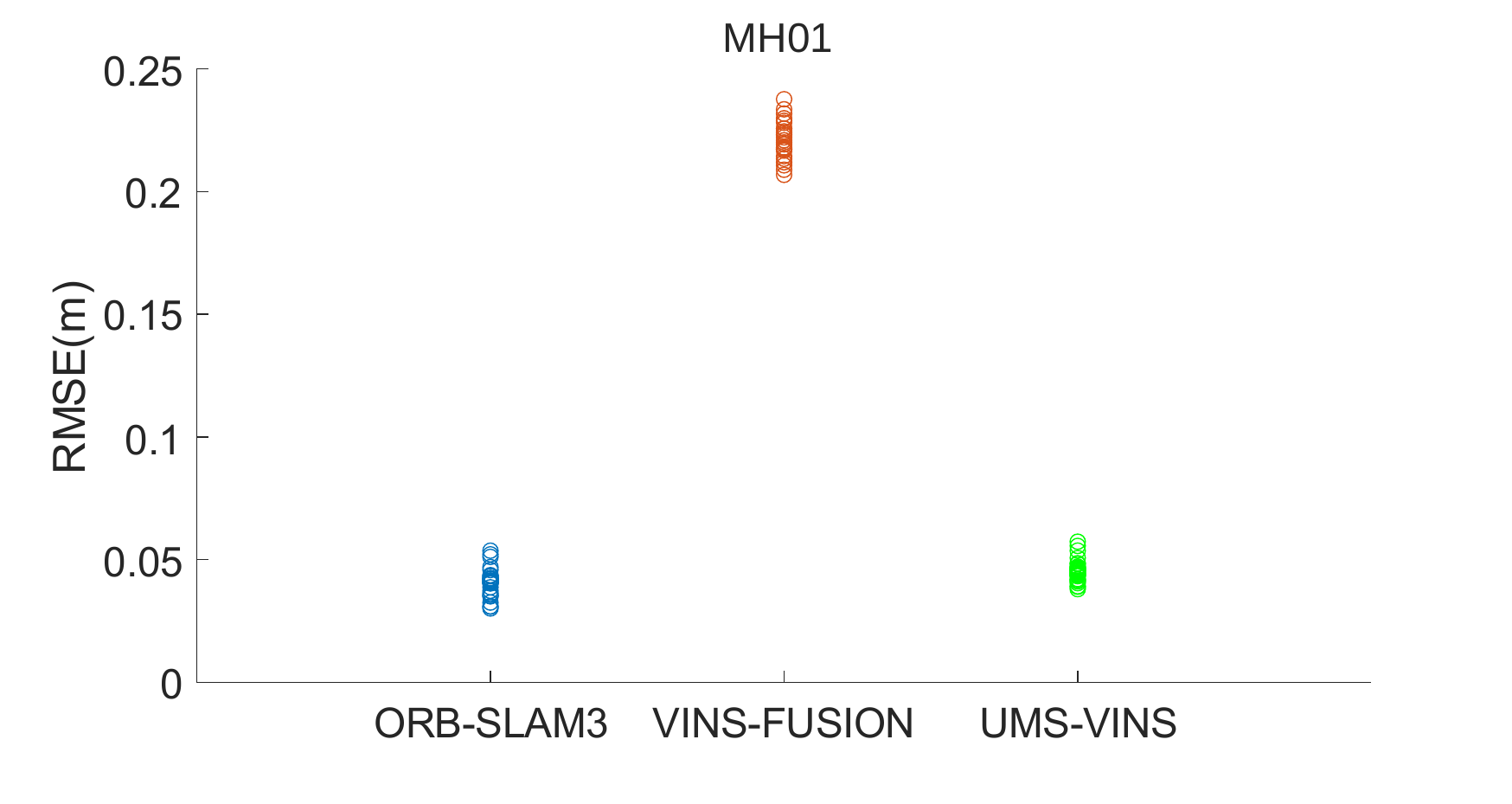}
\caption{}
\end{subfigure}%
\begin{subfigure}[t]{0.53\linewidth}
\centering
\includegraphics[width=1\linewidth]{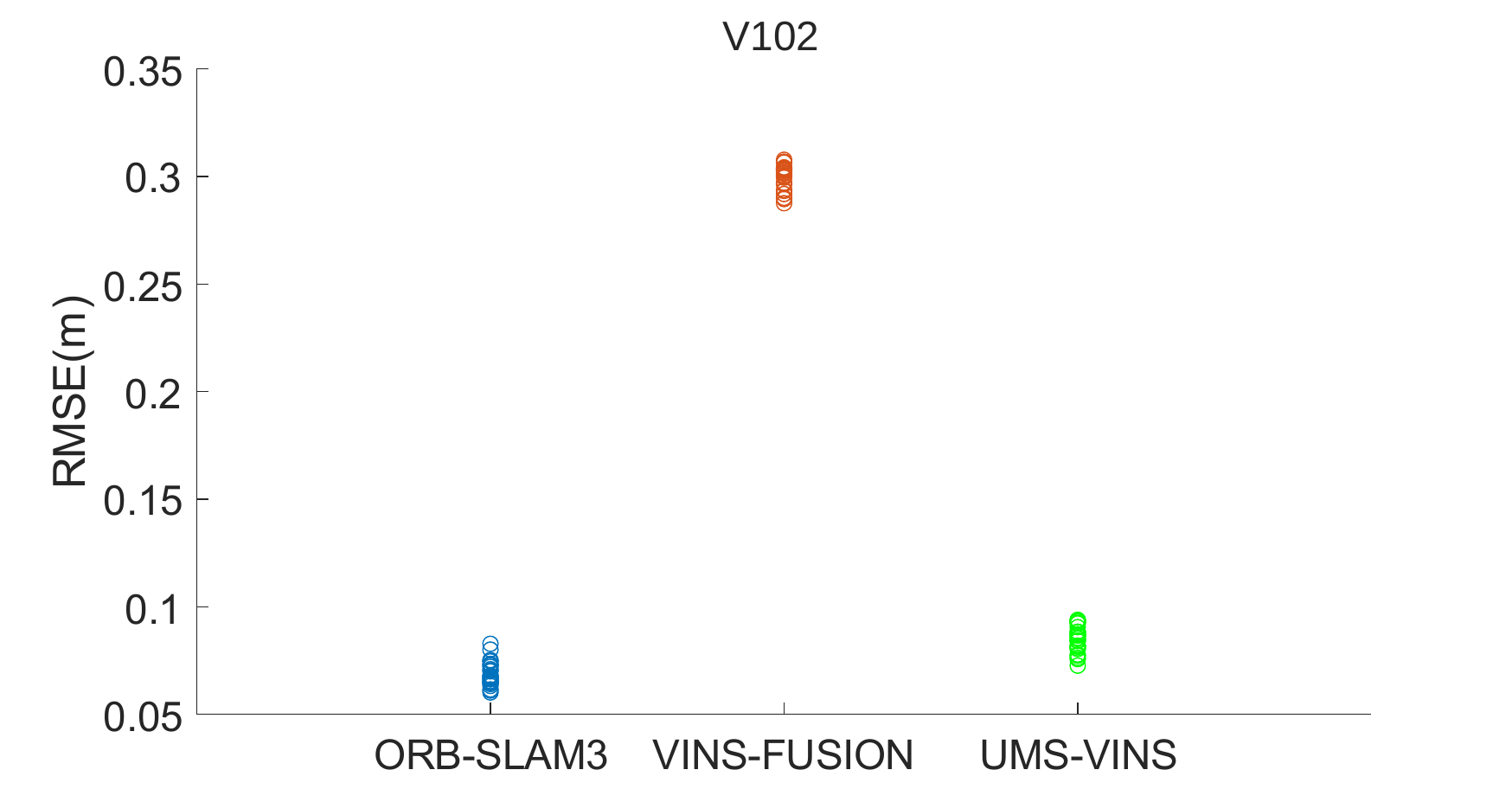}
\caption{}
\end{subfigure}
\caption{The RMSE distribution of  VINS-FUSION, ORB-SLAM3 and UMS-VINS. The three methods run 30 times  for MH\_01 (factory fall) and V1\_02 (small room). The initialization of the 30  runs randomly start from $0\sim 60$ seconds in MH\_01 and V1\_02 sequences. }
\label{Fig:The RMSE variance.}
\end{figure}

To show the initialization robustness of UMS-VINS, we run VINS-FUSION, ORB-SLAM3 and UMS-VINS 30 times on the MH\_01 and the V1\_02 sequences, respectively. For the 30 runs, the initialization randomly start from a moment within $0\sim60$ seconds of the two sequences, as shown in Table \ref{Table-init}. The localization RMSEs of the 30 runs are shown in Fig. \ref{Fig:The RMSE variance.}. The two figures clearly show that the RMSEs of UMS-VINS are similar to those of ORB-SLAM3, but much smaller than those of  UMS-VINS. 

Furthermore, we estimate the variance of the RMSE. For the MH\_01 sequence, the variances of UMS-VINS, VINS-FUSION and ORB-SLAM3 are 0.0021m, 0.0074m and 0.0054m, respectively. The variance of UMS-VINS is smaller than both VINS-FUSION and ORB-SLAM3.  For the V1\_02 sequence, both UMS-VINS and VINS-FUSION obtain 30 results for the 30 runs, respectively, but the ORB-SLAM3 only obtains 26 results and fails in 4 runs because the failure of the initialization. The results of the 30 runs indicate that both UMS-VINS and VINS-FUSION has better environmental adaptability than ORB-SLAM3.   

\subsection{Real-world Experiment}

\begin{figure}[hbt]
\begin{minipage}[t]{0.5\linewidth}
\centering
\includegraphics[width=0.95\linewidth]{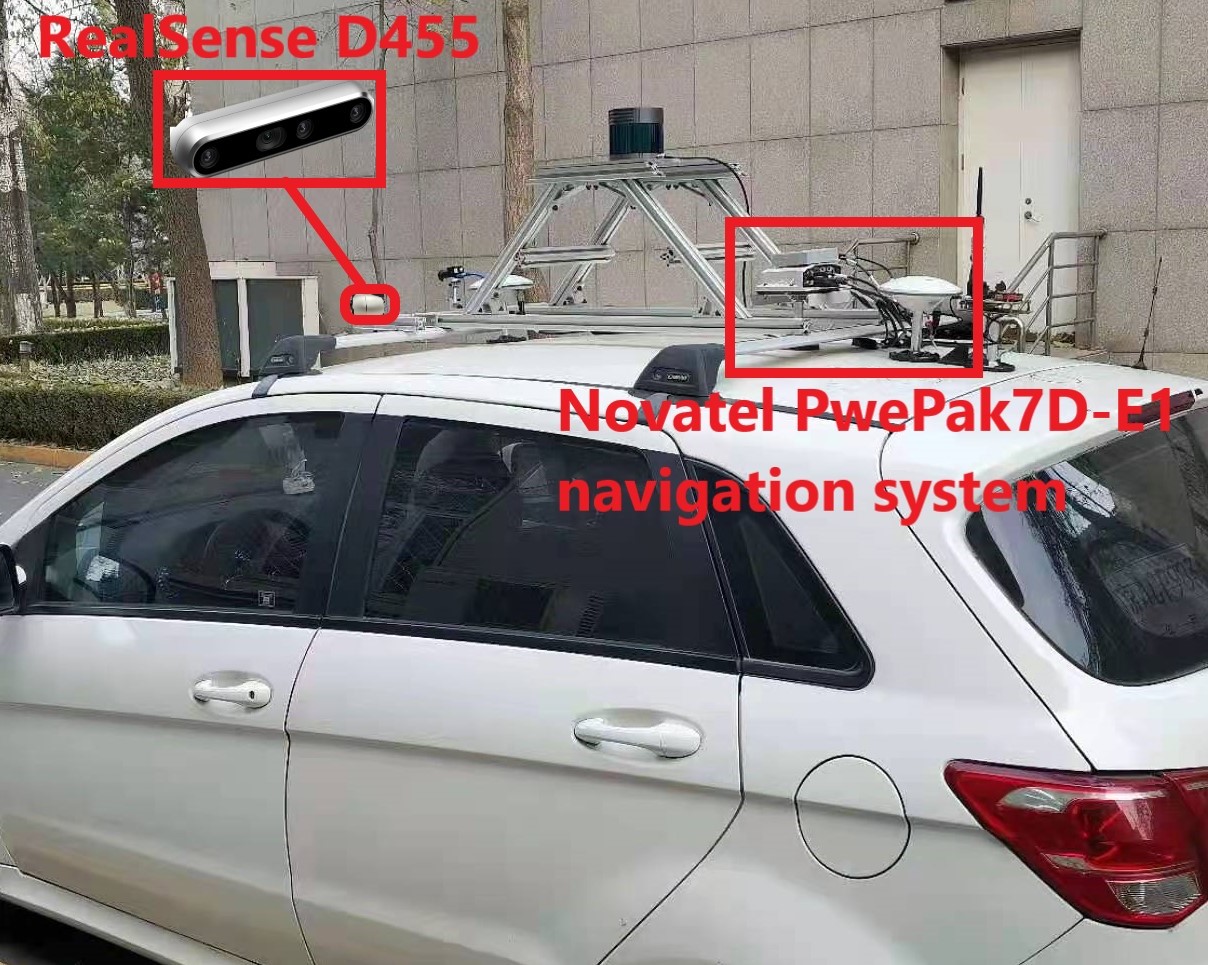}
\caption{The vehicle and sensor layout.}
\label{Fig:The vehicle and the sensor layout.}
\end{minipage}%
\begin{minipage}[t]{0.5\linewidth}
\centering
\includegraphics[width=0.95\linewidth]{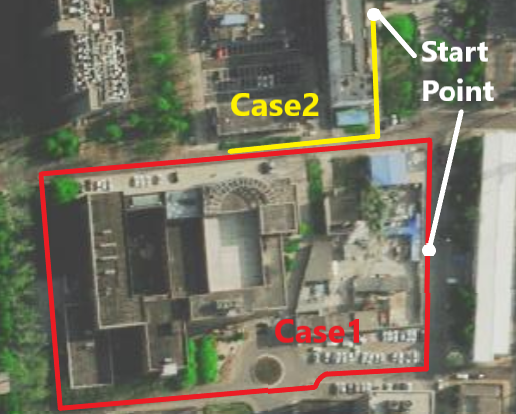}
\caption{The test environment.}
\label{Fig:The test environment.}
\end{minipage}
\end{figure}

We collected the data via an BAIC MOTOR vehicle on Beijing Institute of Technology, Beijing, China. It is an outdoor experiment. A Novatel PwePak7D-E1 navigation system with RTK provides the localization groundtruth. A RealSense D455 consisting of a stereo camera (30FPS, 848x480) and an IMU (100HZ) is used for real-time data collection. The sensor layout and the vehicle are shown in Fig. \ref{Fig:The vehicle and the sensor layout.}.  

Two cases are tested to verify the effectiveness of the proposed UMS-VINS. The test environment and the motion trajectories of the two cases are shown in Fig. \ref{Fig:The test environment.}. Case 1 is in a rich texture environment. The vehicle speed increases from zero. The IMU is well motivated before the experiment. In case 2, the vehicle takes a uniform motion before turning, and the initialization starts from a motion state. The IMU is not motivated. In both cases, the vehicle's straight speed is between 15km/h and 25km/h, and the turning speed is between 5km/h and 10km/h. The radius of the turning circle for case 1 is around 10.0m, and the radius for case 2 is 8.9m. In other words, compared with case 1, case 2 takes a sharp turn. 
\begin{table}[htb]
\caption{Comparison of the localization results in the new real-world experiments (RMSE in m)}
\begin{center}
\begin{tabular}{cccc}
\hline\hline
 & Length(m) & VINS-FUSION & UMS-VINS \\ 
\hline
case 1 & 582.5 & 4.7 & 3.05 \\ 
\hline
case 2 & 60.4 & 33.17 & 0.95 \\
\hline\hline
\end{tabular}
\label{Table-real}
\end{center}
\end{table}

The RMSE of the real-world experimental results can be found in Table \ref{Table-real}. We also run ORB-SLAM3 for both real-world experiments. However, it often suffers from the lost of positioning results due to long-time initialization or positioning drift caused by IMU jitters. The results of ORB-SLAM3 can be found from the link \footnote{https://space.bilibili.com/662047371}. 

\begin{figure}[hbt]
\begin{subfigure}[t]{0.5\linewidth}
\centering
\includegraphics[width=1\linewidth]{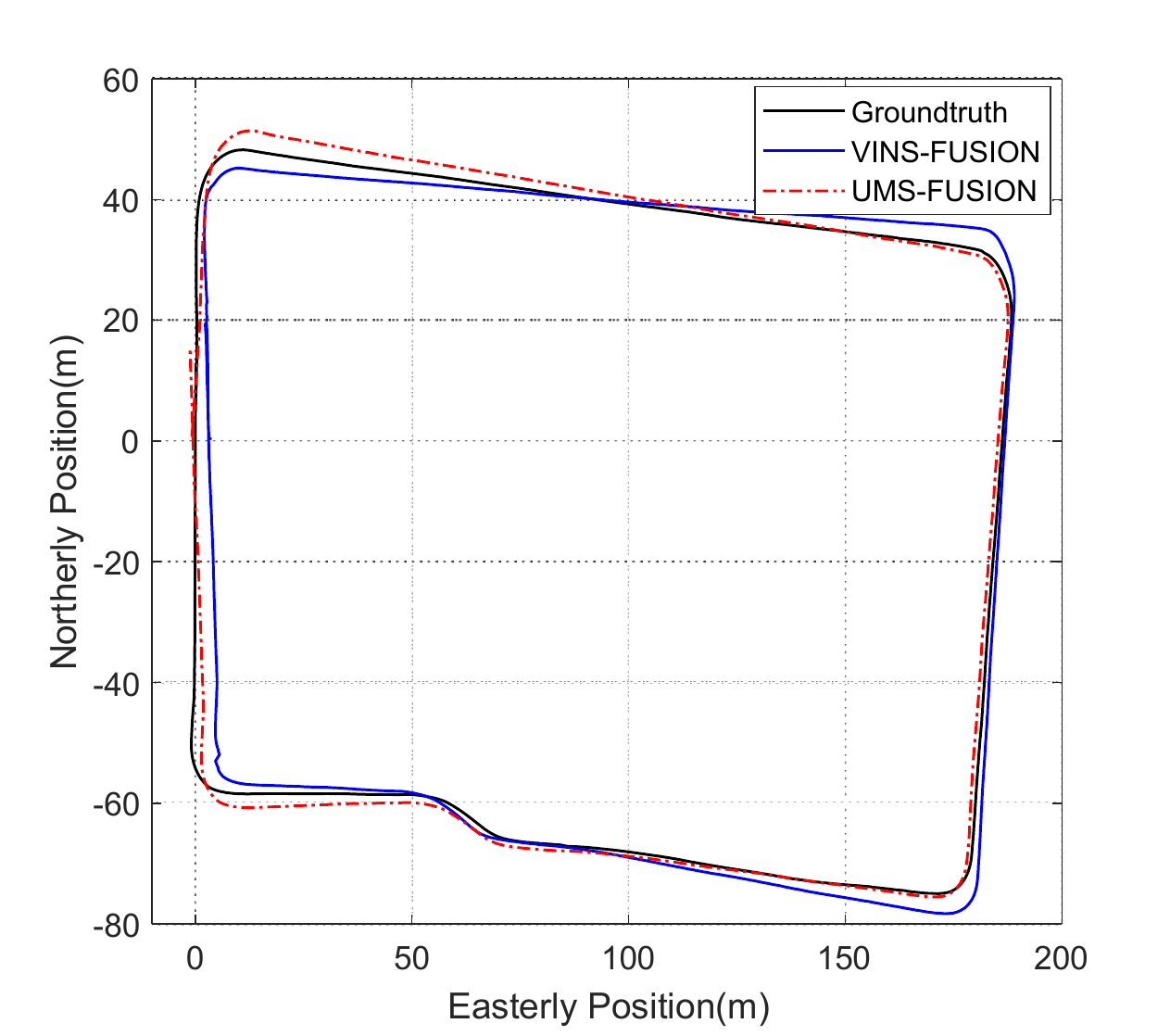}
\caption{Results for case 1.}
\end{subfigure}%
\begin{subfigure}[t]{0.5\linewidth}
\centering
\includegraphics[width=1\linewidth]{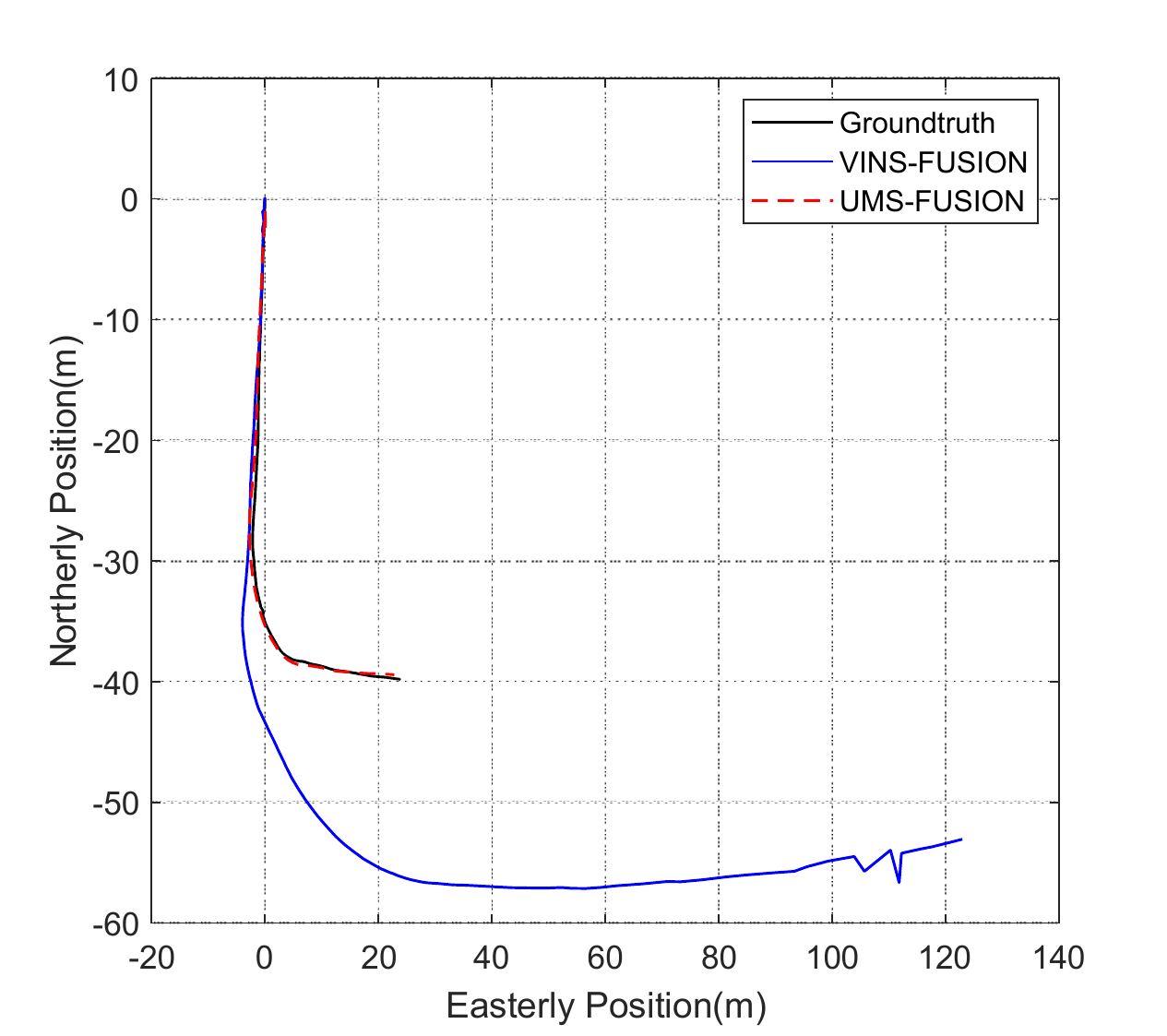}
\caption{Results for case 2.}
\end{subfigure}
\caption{The trajectories estimated from real-world datasets using UMS-VINS and VINS-FUSION.}
\label{results in the real-world.}
\end{figure}

\begin{figure}[hbt]
\begin{subfigure}[t]{0.5\linewidth}
\centering
\includegraphics[width=1\linewidth]{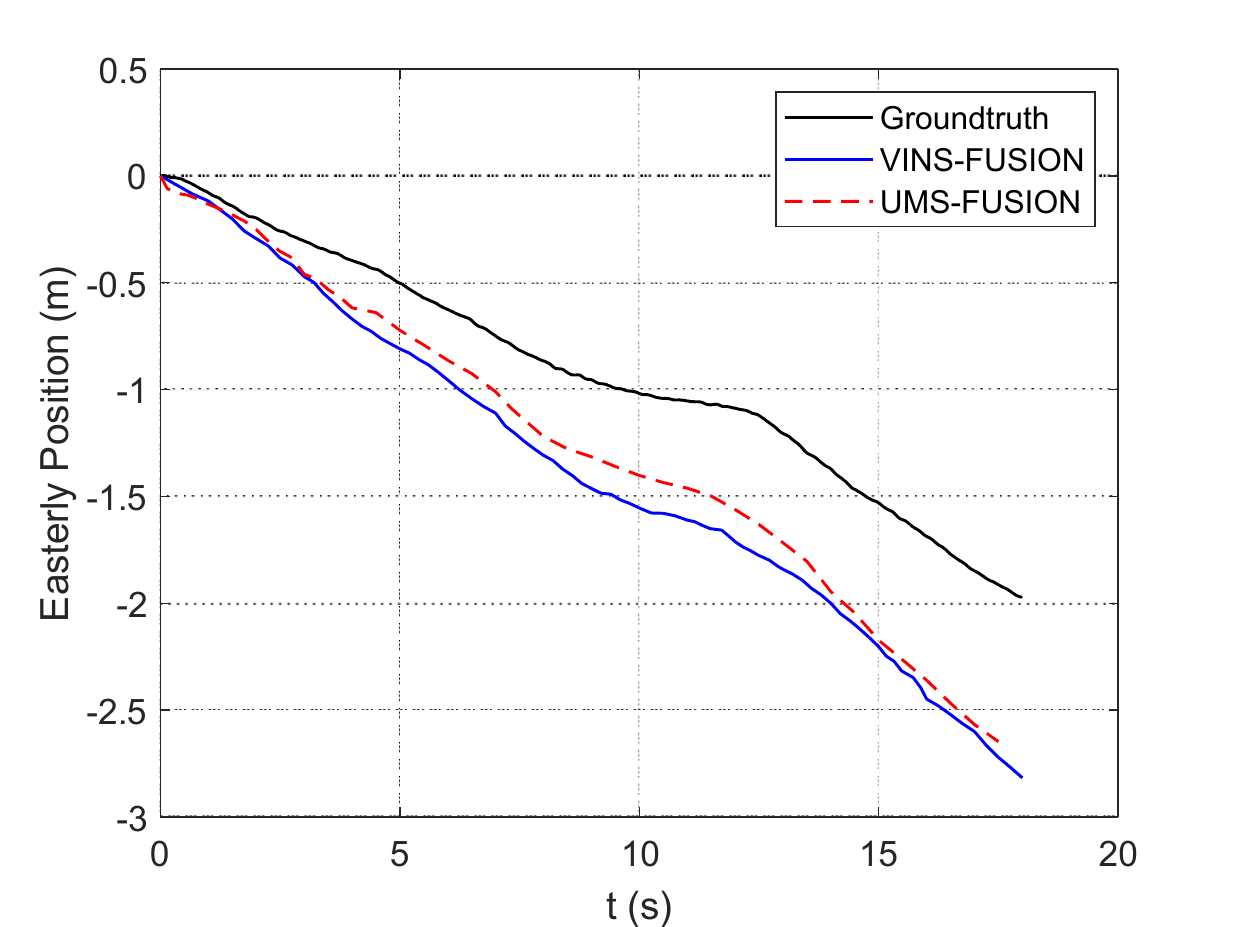}
\caption{Eastern components before turning.}
\end{subfigure}%
\begin{subfigure}[t]{0.5\linewidth}
\centering
\includegraphics[width=1\linewidth]{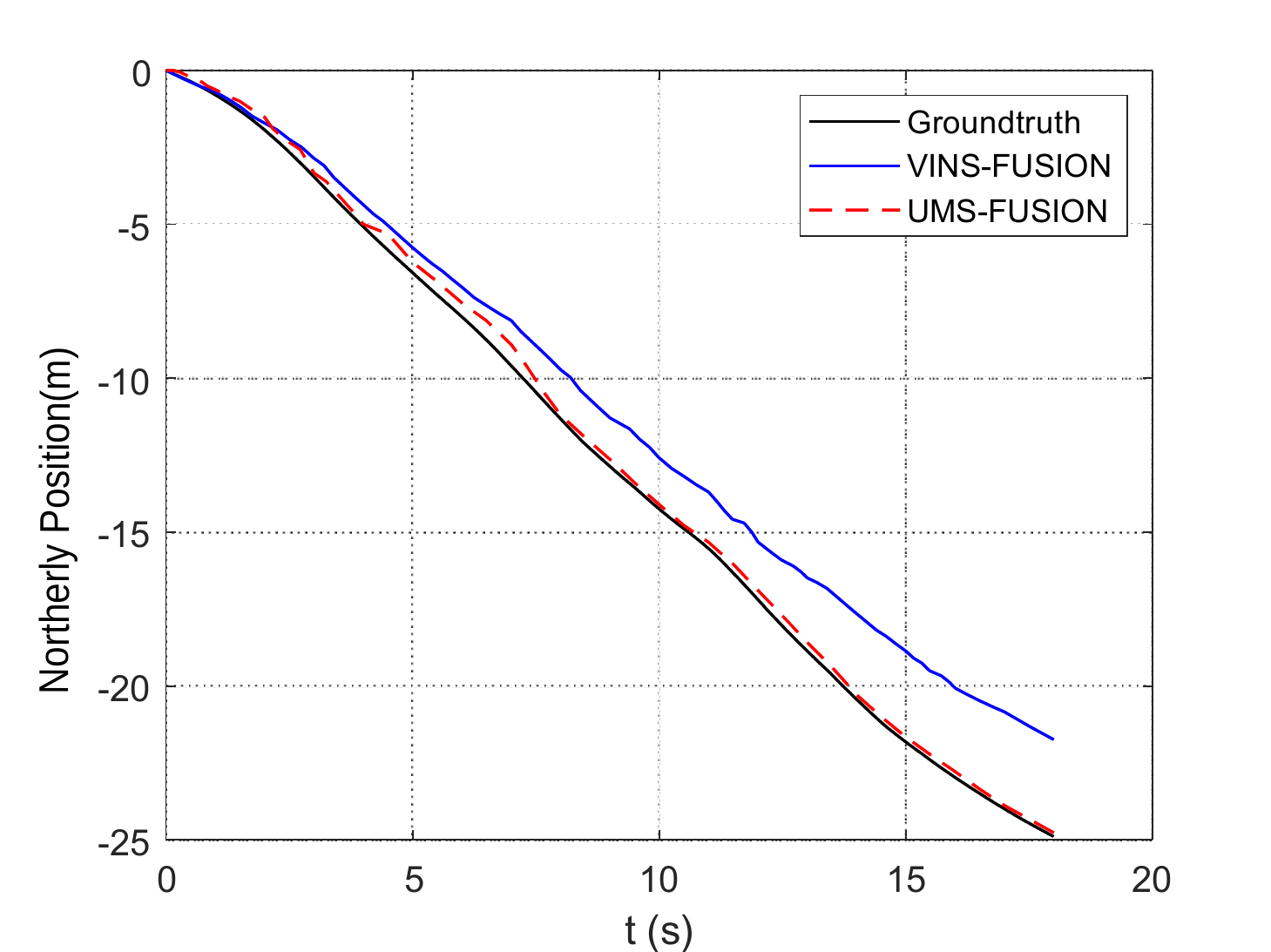}
\caption{Northern components turning.}
\end{subfigure}
\begin{subfigure}[t]{0.5\linewidth}
\centering
\includegraphics[width=1\linewidth]{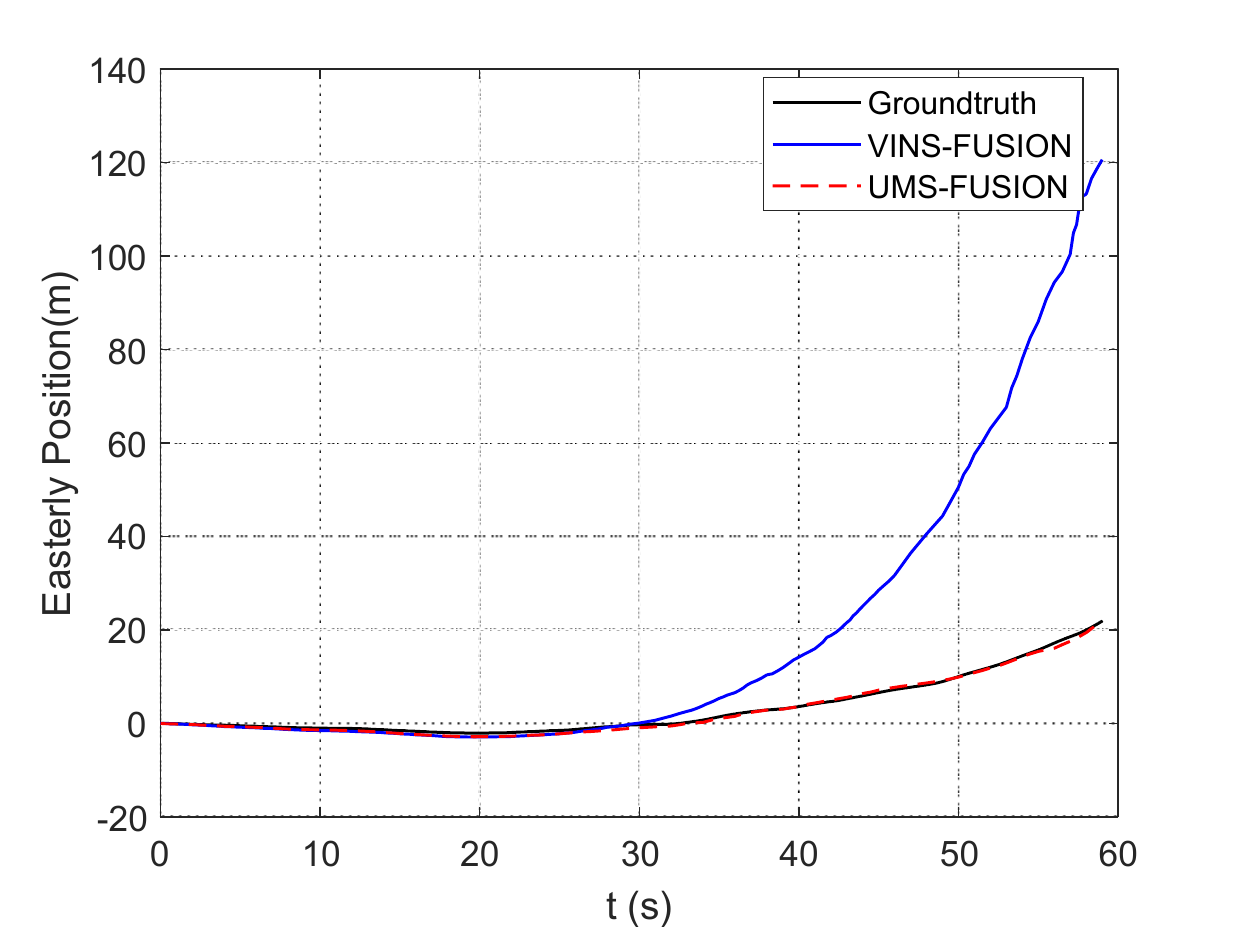}
\caption{Eastern components.}
\end{subfigure}%
\begin{subfigure}[t]{0.5\linewidth}
\centering
\includegraphics[width=1\linewidth]{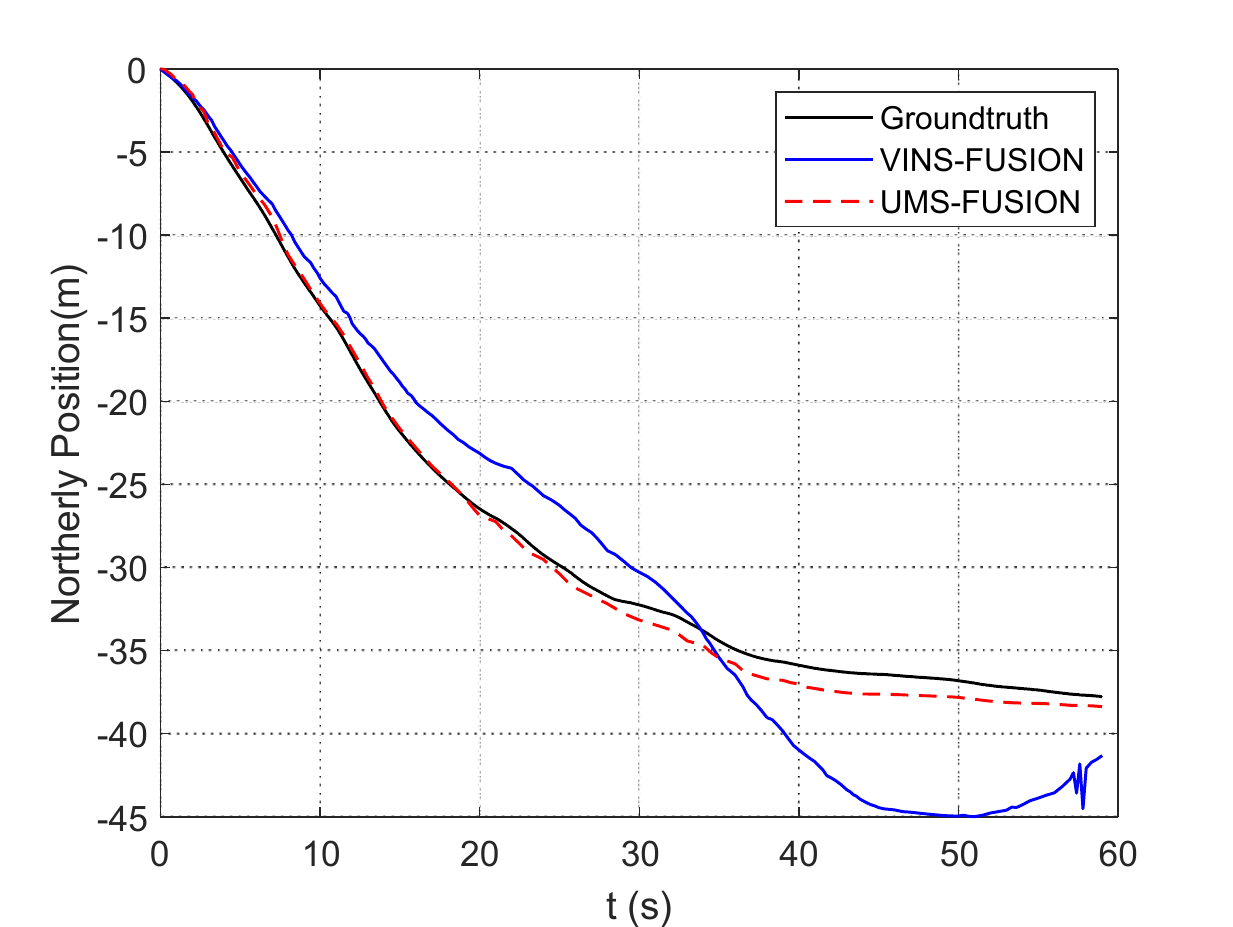}
\caption{Northern components.}
\end{subfigure}
\caption{The northern and eastern components of the positioning results in case 2. }
\label{Fig:The localization results comparison in x and y direction for case 2.}
\end{figure}

For both cases, the proposed UMS-VINS outperforms VINS-FUSION. As shown in Fig. \ref{results in the real-world.}(a), UMS-VINS has a dramatic improvement compared with VINS-FUSION in such an environment-friendly case. 
Both the uniform motion and un-stimulated IMU make case 2 a degeneracy motion. In such a degeneracy case, as shown in Fig. \ref{results in the real-world.}(b), UMS-VINS can work well but VINS-FUSION cannot. To further analyze the failure reason of VINS-FUSION, we decompose the trajectories as can be seen in Fig. \ref{Fig:The localization results comparison in x and y direction for case 2.}.  Fig. \ref{Fig:The localization results comparison in x and y direction for case 2.}(a)-(b) show that before turning, VINS-FUSION can follow the real trajectory, but UMS-VINS outperforms the VINS-FUSION. During the initialization process, the vehicle was uniformly moving. Because of the degree of freedom degeneracy, IMU is not fully motivated, which has a negative influence on the initialization for both UMS-VINS and VINS-FUSION. However, for the proposed UMS-VINS, if the visual-IMU alignment fails, we directly estimate the pose via VO, which can provide better results than the VIO with bad IMU initialization. 

\begin{figure}[hbt]
\begin{subfigure}[t]{0.5\linewidth}
\centering
\includegraphics[width=0.95\linewidth]{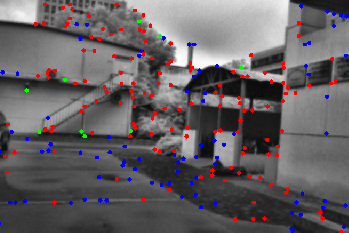}
\caption{UMS-VINS.}
\end{subfigure}%
\begin{subfigure}[t]{0.5\linewidth}
\centering
\includegraphics[width=0.95\linewidth]{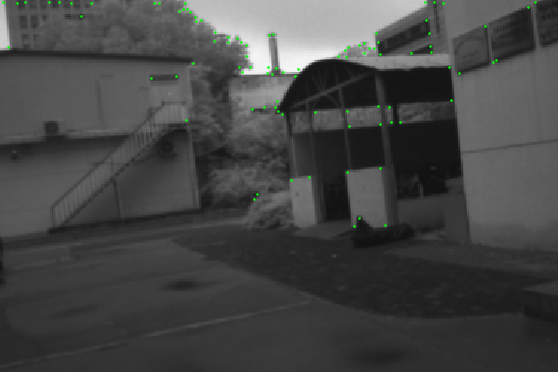}
\caption{VINS-FUSION.}
\end{subfigure}
\caption{The feature extraction during the turning in case 2. In (a), the red points represent 3D features, the blue ones represent 2D features, and the green ones represent features that are not tracked in the current keyframe. The green points in (b) represent the features extracted from VINS-FUSION.}
\label{Fig:The feature extracted in the image.}
\end{figure}

Due to the large yaw variation, the IMU raw observations easily suffer from impulses during the sharp turning, which leads to low positioning performance for VIO.  In UMS-VINS, we add a saturator for the variation of the raw IMU observations, which reduces the influence of the impulses. In addition, as can be seen in Fig. \ref{Fig:The feature extracted in the image.}, during the turning, the number of features extracted by VINS-FUSION is much less than those obtained from UMS-VINS. A large part of the image in Fig. \ref{Fig:The feature extracted in the image.} (b) has no features extracted, which makes the localization worse. The bad IMU initialization, the impulses of IMU raw observations during the turning, the fewer features, and the imbalanced feature distribution make the VINS-FUSION cannot track the real trajectory after the turning, as can be seen in Fig. \ref{Fig:The localization results comparison in x and y direction for case 2.}(c)-(d). On the contrary, the proposed UMS-VINS can well adapt to the environments and follows the real trajectory in case 2. A video of the comparison results can be viewed from link \footnote{https://space.bilibili.com/662047371}.

\section{Conclusions}

The performance of initialization and feature extraction greatly influences the robustness and environmental adaptability of the visual-inertial navigation systems. This paper has developed a visual-inertial tightly coupled odometry by applying the united monocular-stereo features. Both 2D and 3D features are considered, which improve the robustness and environmental adaptability of the visual-inertial navigation systems. The mode switch between VO and VIO reduces the influence of the failure of visual-IMU alignment and guarantees a successful initialization. The positioning performance in public datasets and new real-world experiments indicate that the proposed UMS-VINS outperforms the VINS-FUSION from the perspectives of localization accuracy, localization robustness, and environmental adaptability, and outperforms the ORB-SLAM3 from the perspective of environmental adaptability.

\end{document}